%% file: main.tex
\documentclass[10pt,twocolumn,letterpaper]{article}

\usepackage{mysty}
\usepackage{times}
\usepackage{epsfig}
\usepackage{graphicx}
\usepackage{amsmath}
\usepackage{amssymb}
\usepackage{multirow}
\usepackage{booktabs}
\usepackage{siunitx}
\usepackage{subcaption}
\usepackage{bigstrut}
\usepackage{float}
\usepackage{setspace}

\usepackage{appendix}

\usepackage{tikz}
\usetikzlibrary{decorations.pathmorphing}
\usetikzlibrary{intersections}
\usetikzlibrary{positioning}
\usetikzlibrary{calc}
\usetikzlibrary{fit}
\usepackage[skins]{tcolorbox}

\usepackage{upgreek}
\usepackage{bm}
\makeatletter
\newcommand{\bfgreek}[1]{\bm{\@nameuse{#1}}}
\newcommand{\bfgreekcolor}[2]{\bm{\textcolor{#1}{\@nameuse{#2}}}}
\makeatother

\usepackage{iccv}

\newif\ifshowcomment
\newcommand{\xxcom}[2]{\ifshowcomment\textcolor{#1}{#2}\fi}
\newcommand{\rtcom}[1]{\xxcom{purple}{rt: #1}} %
\newcommand{\ljcom}[1]{\xxcom{green}{lj: #1}} %

\newcommand{\supporapp}[0]{\ifarxiv Appendix\else supplementary material\fi}

\newif\ifarxiv
\arxivtrue

\usepackage[pagebackref=true,breaklinks=true,letterpaper=true,colorlinks,bookmarks=false]{hyperref}

\iccvfinalcopy %

\def\iccvPaperID{739} %

\ificcvfinal\fi
\begin{document}
\title{Context-Aware Zero-Shot Recognition}

\author{Ruotian Luo\\
TTI-Chicago\\
{\tt\small rluo@ttic.edu}
\and
Ning Zhang\\
Vaitl Inc.\\
{\tt\small ning@vaitl.ai}
\and
Bohyung Han\\
Seoul National University\\
{\tt\small bhhan@snu.ac.kr}
\and
Linjie Yang\\
ByteDance AI Lab\\
{\tt\small linjie.yang@bytedance.com}
}
\maketitle

\input{sections/abstract_intro}

\input{sections/related}

\input{sections/models}

\input{sections/experiments}

\input{sections/conclusion}

\ificcvfinal
\fi

{\small
\bibliographystyle{ieee}
\bibliography{egbib}
}

\ifarxiv
\newpage
\appendix
\appendixpage
\input{sections/appendix}

\fi
\end{document}

%% file: sections/abstract_intro.tex
\begin{abstract}

We present a novel problem setting in zero-shot learning, zero-shot object recognition and detection in the context.
Contrary to the traditional zero-shot learning methods, which simply infers unseen categories by transferring knowledge from the objects belonging to semantically similar seen categories, we aim to understand the identity of the novel objects in an image surrounded by the known objects using the inter-object relation prior.
Specifically, we leverage the visual context and the geometric relationships between all pairs of objects in a single image, and capture the information useful to infer unseen categories. 
We integrate our context-aware zero-shot learning framework into the traditional zero-shot learning techniques seamlessly using a Conditional Random Field (CRF). 
The proposed algorithm is evaluated on both zero-shot region classification and zero-shot detection tasks. 
The results on Visual Genome (VG) dataset show that our model significantly boosts performance with the additional visual context compared to traditional methods.
\ificcvfinal
The code is available at \url{https://github.com/ruotianluo/Context-aware-ZSR}.
\fi

\end{abstract}

\section{Introduction}\label{sec:intro}

Supervised object recognition has achieved substantial performance improvement thanks to the advance of deep convolutional neural networks in the last few years~\cite{ren2015faster,redmon2017yolo9000,Hu_2018_CVPR,girshick2015fast}. 
Large-scale datasets with comprehensive annotations, {\it e.g.}, COCO~\cite{lin2014microsoft}, facilitate deep neural networks to learn semantic knowledge of the objects within a predefined set of classes. 
However, it is impractical to obtain rich annotations for every class in the world while it is important to develop the models that can generalize to new categories without extra annotations.
On the other hand, human beings have capability to understand the unseen object categories using external knowledge such as language descriptions and object relationships. 
The problem of inferring objects in unseen categories is referred to as \emph{zero-shot} object recognition in recent literature~\cite{fu2018recent,xian2018zero}.

In the absence of direct supervision, other resources of information such as semantic embedding~\cite{norouzi2013zero}, knowledge graph~\cite{wang2018zero,rohrbach2011evaluating}, and attributes~\cite{sung2018learning,changpinyo2016synthesized} are often employed to infer the appearance of novel object categories through knowledge transfer from seen categories.
The assumption behind the approaches is that if an unseen category is semantically close to a seen category, objects of the two categories should be visually similar.

\begin{figure}[t]
\centering
\includegraphics[width=\linewidth]{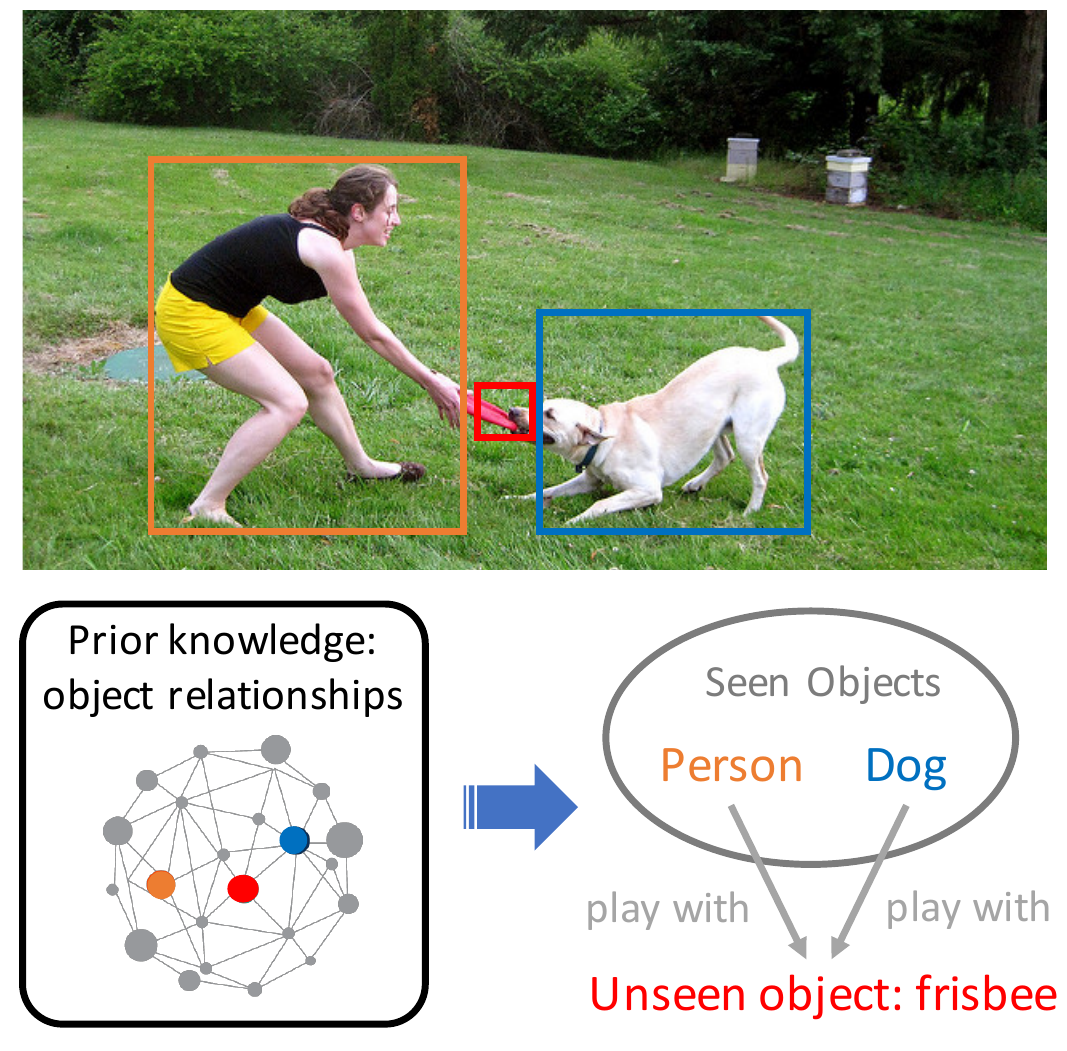}
\caption{An example of zero-shot recognition with context information. It contains two seen objects (person and dog) and one unseen object (frisbee). The prior knowledge of relationships between seen and unseen categories provide cues to resolve the category of the unseen object.}
\label{figs:context}
\end{figure}

Besides inferring novel object categories using visual similarity, human often capture the information of an object in the scene context. 
For example, if we do not know the class label of the red disk-like object in the middle of the image shown in Figure~\ref{figs:context}, it is possible to guess its category even with the limited visual cues by recognizing two other objects in the neighborhood, a person and a dog, and using the prior knowledge about the objects that a person and a dog potentially play with together. 
Suppose that a frisbee is known to be such kind of an object, we can infer the object as a frisbee even without seeing it before. 
In this scenario, the interaction between multiple objects, {\it e.g.}. person, dog, and frisbee, provides additional clues to recognize the novel object---frisbee in this case; note that the external knowledge about the object relationships (person and dog can play with frisbee) is required for unseen object recognition. 

Motivated by this intuition, we propose an algorithm for zero-shot image recognition \emph{in the context}. 
Different from the traditional methods that infer each of unseen objects independently, we aim to recognize novel objects in the visual context, {\it i.e.}, by leveraging the relationships of the objects shown in an image. 
The relationship information is defined by a relationship knowledge graph in our framework and it is more straightforward to construct a knowledge graph than to collect dense annotations on images. 
In our framework, a Conditional Random Field (CRF) is employed to jointly reason over local context information as well as relationship graph prior.
Our algorithm is evaluated on Visual Genome dataset~\cite{krishna2017visual}, which provides a large number of object categories and diverse object relations; our model based on the proposed context knowledge representation illustrates the clear advantage when applied to various existing methods for zero-shot recognition. 
We believe the proposed topic will foster more interesting work in the domain of zero-shot recognition.

The main contributions of this work are as follows:
\begin{itemize}
    \item We introduce a new framework of zero-shot learning in computer vision, referred to as zero-shot recognition in the context, where unseen object classes are identified by the relation to other ones shown in the same image.%
    \vspace{-.5em}
    \item We propose a novel model based on deep neural networks and CRF, which learns to leverage object relationship knowledge to recognize unseen object classes.%
    \vspace{-.5em}
    \item The proposed algorithm achieves the significant improvement compared to existing methods on various models and setting that ignore visual context.  
\end{itemize}

The rest of the paper has the following organization.
Section~\ref{sec:related} review existing zero-shot learning techniques for visual recognition.
Section~\ref{sec:context} and \ref{sec:implementation} describes our main algorithm and its implementation details, respectively.
Experimental results are discussed in Section~\ref{sec:exp} and we conclude the paper in Section~\ref{sec:conclusion}.

%% file: sections/related.tex
\section{Related work}\label{sec:related}
This section present the prior works related to our work including zero-shot learning, context-aware recognition, and knowledge graph.

\subsection{Zero-shot learning} 
A wide range of external knowledge has been explored for zero-shot learning. 
Early zero-shot classification approaches adopt object attributes as a proxy to learn visual representation of unseen categories~\cite{sung2018learning,changpinyo2016synthesized,changpinyo2018classifier}. 
Semantic embeddings are learned from large text corpus and then utilized to bridge seen and unseen categories~\cite{frome2013devise,norouzi2013zero}. 
Combination of attributes and word embeddings are employed to learn classifiers of unseen categories by taking linear combinations of synthetic base classifiers~\cite{changpinyo2016synthesized,changpinyo2018classifier}, and text descriptions are also incorporated later to predict classifier weights~\cite{lei2015predicting}. 
A recent work~\cite{wang2018zero,kampffmeyer2018rethinking} applies Graph Convolutional Network (GCN)~\cite{duvenaud2015convolutional} over WordNet knowledge graph to propagate classifier weights from seen to unseen categories. 
More detailed survey can be found in \cite{fu2018recent,xian2018zero}. 

In addition to these knowledge resources, we propose to exploit the object relationship knowledge in the visual context to infer unseen categories. 
To the best of our knowledge, this is the first work to consider pairwise object relations for zero-shot visual recognition. 
The proposed module can be easily incorporated into existing zero-shot image classification models, leading to performance improvement.

In addition to zero-shot recognition, zero-shot object detection (ZSD) task is also studied, which aims to localize individual objects of categories that are never seen during training~\cite{Bansal_2018_ECCV,Tao_2018_ECCV,rahman2018zero,zhu2018zero,demirel2018zero}. 
Among the approaches, \cite{zhu2018zero} focuses on generating object proposals for unseen categories while \cite{Bansal_2018_ECCV} trains a background-aware detector to alleviate the corruption of the ``background'' class with unseen classes. 
Also, \cite{rahman2018zero} proposes a novel loss function to reduce noise in semantic features. 
Although these methods handle object classification and localization jointly, none of them have attempted to incorporate context information in the scene.

\subsection{Context-aware detection}
Context information has been used to assist object detection before deep learning era \cite{galleguillos2010context,divvala2009empirical,felzenszwalb2010object,galleguillos2008object,desai2011discriminative}. 
Deep learning approaches such as Faster R-CNN~\cite{ren2015faster} allow a region feature to look beyond its own bounding box via the large receptive field. 
Object relationships and visual context are also utilized to improve object detection. For example, \cite{Yang_2018_ECCV,li2017msdn} show that the joint learning of scene graph generation and object detection improves detection results while \cite{Chen_2018_ECCV,Hu_2018_CVPR} perform message passing between object proposals to refine detection results. 
A common-sense knowledge graph is used for weakly-supervised object detection~\cite{Singh_2018_ECCV}. 
For the categories without localization annotations, the common-sense knowledge graph is employed to infer their locations, which are then used as training data.

Although context-aware methods have been studied for object detection for a while, these methods are mostly designed for fully-supervised setting thus cannot be directly applied to zero-shot environment. 
For example, \cite{divvala2009empirical} uses occurrence frequency of object pairs, which is not available for unseen categories. \cite{Yang_2018_ECCV} uses densely annotated scene graphs of all object categories to improve detection accuracy. In this paper, we explore to port context-aware idea to zero-shot setting.

\subsection{Knowledge graphs}
Knowledge graphs has been applied to various vision tasks including image classification~\cite{marino2016more,lee2017multi}, zero-shot learning~\cite{rohrbach2010helps,rohrbach2011evaluating,wang2018zero}, visual reasoning~\cite{malisiewicz2009beyond,zhu2014reasoning,Chen_2018_CVPR}, and visual navigation~\cite{yang2018visual}.
Graph-based neural networks often propagate information on the knowledge graph~\cite{marino2016more,lee2017multi,wang2018zero,Chen_2018_CVPR}. Following \cite{marino2016more,Chen_2018_CVPR,yang2018visual}, we construct the relationship knowledge graph used in our method in a similar way.%

%% file: sections/models.tex
\begin{figure*}[t]
\centering
\includegraphics[width=\textwidth]{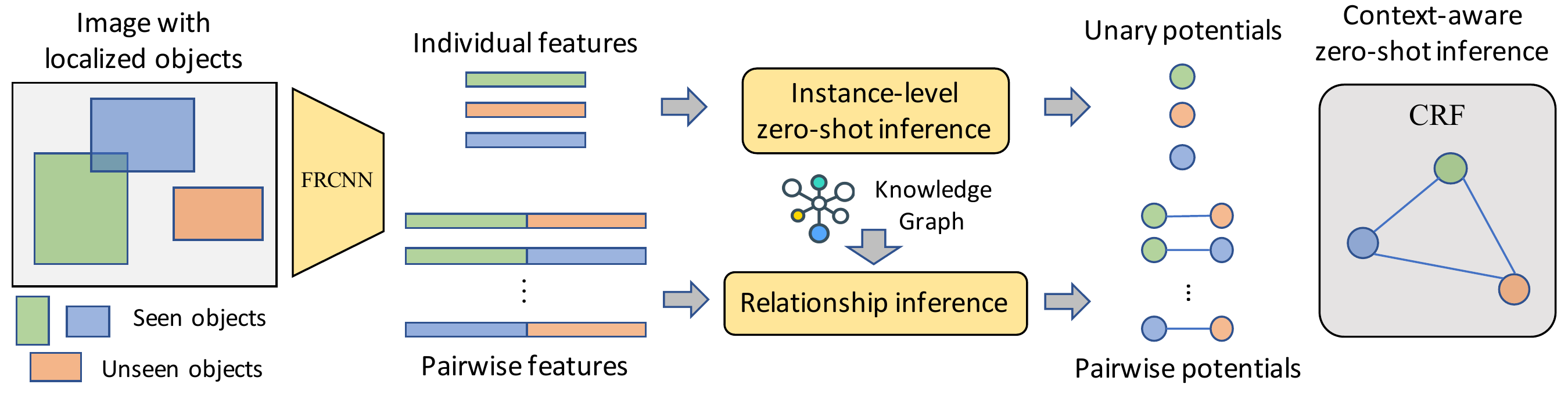}
\caption{The overall pipeline of our algorithm. First, features for individual objects as well as object pairs are extracted from the image. An instance-level zero-shot inference module is applied on individual features to generate unary potentials. A relationship inference module takes pairwise features and relationship knowledge graph to generate pairwise potentials. Finally, the most likely object labels are inferred from CRF constructed by generated potentials.}
\label{figs:model}
\end{figure*}

\section{Context-aware zero-shot recognition}\label{sec:context} 
\subsection{Problem formulation}
The existing zero-shot recognition techniques~\cite{frome2013devise,lampert2014attribute} mostly focus on classifying objects independently with no consideration of potentially interacting objects.
To facilitate context-aware inference for zero-shot recognition, we propose to classify all the object instances---both seen and unseen objects---in an image.
We first assume that the ground-truth bounding box annotations are given and propose to recognize objects in the unseen classes.
After that, we also discuss zero-shot object detection when the ground-truth bounding boxes are not available at test time. 

Our model takes an image $I$ and a set of bounding boxes (regions) $\{B_i\}$ as its inputs, and produeces a class label $c_i$ out of the label set $\mathcal{C}$ for each region. 
Under the zero-shot recognition setting, the label set $\mathcal C$ is split into two subsets, $\mathcal S$ for seen categories and $\mathcal U$ for unseen categories, where the two sets satisfy $\mathcal S \cup \mathcal U=  \mathcal C$ and $\mathcal S \cap \mathcal U = \varnothing$.  
The object labels in $\mathcal S$ are available during training while the ones in $\mathcal U$ are not.
The model needs to classify regions of both seen and unseen categories in testing.

Some existing zero-shot recognition approaches have utilized knowledge graph~\cite{wang2018zero} for transfer learning from seen to unseen categories, where an object in an unseen category is recognized through the cues from the related seen categories in the knowledge graph. 
The edges in the knowledge graph typically represent visual similarity or hierarchy. 
In our formulation, a relationship knowledge graph has edges representing the ordered pairwise relationships in the form of {$<$subject, predicate, object$>$}, which indicate the possible interactions between a pair of objects in an image.
A directed edge denotes a specific predicate (relation) in the relationship given by a tuple {$<$subject, predicate, object$>$}. 
We may have multiple relations for the same pair of categories; in other words, there can be multiple relationships defined on an ordered pair of categories. 
Given a set of relations, $\mathcal{R} = \{ r_k | k = 1, \dots, K \}$, the relationship graph is defined by $\textsf{G} = \{ \mathcal{V}, \mathcal{E} \}$, where $\mathcal{V}$ denotes a set of classes and $\mathcal{E} = \{r_{mn}^{(i)} \in \mathcal{R} | i = 1, \dots, K_{mn} \text{ and } m, n \in \mathcal{C} \} $ is a set of directed edges representing relations between all pairs of a subject class $m$ and an object class $n$.
Note that $K_{mn}$ is the number of all possible predicates between the ordered pair of classes.

\subsection{Our framework}
Our framework is illustrated in Figure~\ref{figs:model}. 
From an image with localized objects, we first extract features from the individual objects and the ordered object pairs. 
We then apply an instance-level zero-shot inference module to the individual object features, and obtain a probability distribution of the object over all object categories. 
The individual class likelihoods are used as unary potentials in the unified CRF model. 
A relationship inference module takes the pairwise features as an input and computes the corresponding pairwise potentials using the relationship graph. 

Specifically, let $B_i$ and $c_i$ ($i = 1,\ldots,N$) be an image region and a class assignment of $N$ objects in an image. 
Our CRF inference model is given by
\begin{align}
    P&(c_1\ldots c_N|B_1\ldots B_N) \nonumber \\
    & \propto \exp \left( \sum_{i}\theta(c_i|B_i)+\gamma\sum_{i\neq j}\phi(c_i, c_j | B_i, B_j) \right)
\label{eq:pgm}
\end{align}
where the unary potential $\theta(c_i|B_i)$ comes from the instance-level zero-shot inference module and the pairwise potential $\phi(c_i, c_j | B_i, B_j)$ is obtained from the relationship inference module. $\gamma$ is a weight parameter balancing between unary and pairwise potentials.

The final prediction is generated through the MAP inference on the CRF model given by Eq.~\eqref{eq:pgm}. 
We call the whole procedure \emph{context-aware zero-shot inference}. 
Similar techniques can be found in context-aware object detection techniques~\cite{divvala2009empirical,galleguillos2010context}.
However, we claim that our algorithm has sufficient novelty because we introduce a new framework of zero-shot learning with context and design the unary and pairwise potentials specialized in CRF for zero-shot setting.
We hereafter use $\theta_i(\cdot)$ and $\phi_{ij}(\cdot)$ as the abbreviations for $\theta(\cdot|B_i)$ and $\phi(\cdot| B_i, B_j)$, respectively.
We discuss the detail of each component in the CRF next.

\subsubsection{Instance-level zero-shot inference}\label{sec:instance-level}

We use a modified version of Fast R-CNN framework~\cite{girshick2015fast} to extract features from individual objects. 
The input image and the bounding boxes are passed through a network composed of convolutional layers and RoiAlign~\cite{he2017mask} layer. 
The network outputs a region feature $\mathbf{f}_i \in \mathbb{R}^{d_\mathbf{f}}$ for each region, which is further forwarded to a fully connected layer to produce the probability of each class $P_{\text{c}}(c_i) = \text{softmax}(\mathbf{W} \mathbf{f}_i)$, where $\mathbf{W} \in \mathbb{R}^{|\mathcal{C}| \times d_{\mathbf{f}}}$ is a weight matrix. 
The unary potential of the CRF is then given by
\begin{align}
    \theta_i(c_i) = \log P_{c}(c_i|B_i)
\end{align}
Although it is straightforward to learn the network parameters including $\mathbf{W}$ in the fully supervised setting, we can train the model only for the seen categories and obtain $\mathbf{W}_\mathcal S \in \mathbb{R}^{|\mathcal{S}| \times d_\mathbf{f}}$. 
To handle the classification of unseen category objects, we have to estimate $\mathbf{W}_\mathcal{U}$ as well and construct the full parameter matrix $\mathbf{W} = [\mathbf{W}_{\mathcal S}^\top, \mathbf{W}_{\mathcal U}^\top]^\top$ for prediction.
There are several existing approaches~\cite{changpinyo2016synthesized,lei2015predicting,changpinyo2017predicting} to estimate the parameters for the unseen categories from external knowledge. 
We will evaluate the performance of our context-aware zero-shot learning algorithm in the several parameter estimation techniques for unseen categories in Section~\ref{sec:exp}.

\subsubsection{Relationship inference with relationship graph}

The pairwise potential of the CRF model is given by a relationship inference module.
It takes a pair of regions as its inputs and produces a relation potential, $\ell(\hat{r}_k ; B_i,B_j)$, which indicates the likelihood of the relation $\hat{r}_k$ between the two bounding boxes.
Then the pairwise potential of the CRF is formulated as
\begin{align}
\phi(c_i, c_j | B_i, B_j) = \sum_k {\delta(\hat{r}_k; c_i, c_j) \ell(\hat{r}_k ; B_i, B_j)},
\label{eq:pairwise}
\end{align}
where $\delta(\hat{r}_k ; c_i, c_j) $ is an indicator function whether tuple $<$$c_i, \hat{r}_k, c_j$$>$ exists in the relationship graph.
Intuitively, a label assignment is encouraged when the possible relations between the labels have large likelihoods.

The relationship inference module estimates the pairwise potential from a geometric configuration feature using an embedding function followed by a two-layer multilayer perceptron as
\begin{align}
    \ell(r | B_i,B_j) = \text{MLP}(t_{\eta}(g_{ij})),
    \label{eq:relationship}
\end{align}
where $g_{ij}$ is the relative geometry configuration feature of two objects corresponding to $B_i$ and $B_j$ based on \cite{Hu_2018_CVPR} and $t_\eta(\cdot)$ embeds its input onto a high-dimensional space by computing cosine and sine functions of different wavelengths~\cite{vaswani2017attention}.
Formally, translation- and scale-invariant feature $g_{ij}$ is given by 
\begin{equation}
\hspace{-0.1cm} g_{ij} = \left[ \log \frac{|x_i-x_j|}{w_i}, \log \frac{|y_i-y_j|}{h_i}, \log \frac{w_j}{w_i}, \log \frac{h_j}{h_i} \right]^\top \hspace{-0.2cm},
\end{equation}
where $(x_i, y_i, w_i, h_i)$ represents the location and size of $B_i$.

To train the MLP in Eq.~\eqref{eq:relationship}, we design a loss function based on pseudo-likelihood, which is the likelihood of a region given the ground-truth labels of the other regions. 
Maximizing the likelihood increases the potential of true label pairs while suppressing the wrong ones.
Let $c_i^*$ to be the ground-truth label of $B_i$. 
The training objective is to minimize the following loss function:
\begin{equation}
    L = - \sum_i \log P(c_i^*|c_{\backslash i}^*),
\end{equation}
where $c_{\backslash i}^*$ denotes the ground-truth labels of bounding boxes other than $B_i$ and
\begin{align}
    P&(c_i^*|c_{\backslash i}^*)  \label{eq:loss}\\
    &= \frac{ \exp \sum_{j\neq i}[\theta_i(c^*_i)+\gamma\phi_{ij}(c^*_i, c^*_j)+ \gamma\phi_{ji}(c^*_j, c^*_i)]}{\sum_{c\in\mathcal S} \exp \sum_{j\neq i}[\theta_i(c)+\gamma\phi_{ij}(c, c^*_j) + \gamma\phi_{ji}(c^*_j, c)]}.\nonumber
\end{align}
Note that $\ell(r | B_i, B_j)$ is learned implicitly through optimizing of this loss. 
No ground-truth annotation about relationships is used in training.

\subsubsection{Context-aware zero-shot inference} \label{sec:context-aware}
The final step is to find the assignment that maximizes $P(c_1, \dots, c_N)$ given the trained CRF defined by Eq.~\eqref{eq:pgm}. 
We adopt mean field inference \cite{koller2009probabilistic} for efficient approximation. 
A distribution $Q(c_1, \dots, c_N)$ is used to approximate $P(c_1, \dots, c_N)$, which is given by the product of the independent marginals, which is given by
\begin{equation}
Q(c_1, \dots, c_N) = \prod_{i}Q_i(c_i)
\end{equation}
To get a good approximation of $Q$, we minimize the KL-divergence, $\text{KL}(Q\|P)$, while constraining $Q(c_1, \dots, c_N)$ and $Q_i(c_i)$ to be valid distributions. 
The optimal $Q$ is obtained by iteratively updating $Q$ using the following rule:
\begin{align}
\hspace{-0.23cm} Q_i(c_i) \hspace{-0.05cm} \leftarrow  \hspace{-0.05cm} \frac{1}{Z_i}  \hspace{-0.05cm} \exp \hspace{-0.1cm} \left( \hspace{-0.05cm} \theta_i(c_i) +
\gamma\sum_{j\neq i} \hspace{-0.05cm} \sum_{c_j\in \mathcal C}Q_j(c_j)\phi_{ij}(c_i,c_j) \hspace{-0.05cm} \right)  \hspace{-0.1cm} ,
\end{align}
where $Z_i$ is a partition function.

The pairwise potential defined in Eq.~\eqref{eq:pairwise} involves a $(N \times |\mathcal C|)^2\times |\mathcal R|$ matrix. 
Since it may incur a huge computation overhead when $N$ and $|\mathcal C|$ are large, we perform pruning for acceleration.
We select the categories with top $K$ probabilities in terms of $P_{\text{c}}$. 
In this way, our method can be viewed as a cascade algorithm; the instance-level inference serves as the first layer of the cascade, and the context-aware inference refines the results using relationship information.

%% file: sections/experiments.tex
\section{Implementation}
\label{sec:implementation}

This section discusses more implementation-oriented details of our zero-shot recognition algorithm.

\subsection{Knowledge graph} 
We extract our relationship knowledge graph from Visual Genome dataset, similar to~\cite{marino2016more,Chen_2018_CVPR,yang2018visual}.
We first select 20 most frequent relations and collect all the subject-object relationships that (1) occurs more than 20 times in the dataset and (2) have the relation defined in $\mathcal R$.
The purpose of this process is to obtain a knowledge graph with common relationships. 
The relation set $\mathcal R$ includes `on', `in', `holding', `wearing' etc. 
\ificcvfinal
\else
We will release our code, pretrained model and this relationship knowledge graph once the paper is accepted.
\fi

\subsection{Model}
We build our model based on a PyTorch Mask/Faster R-CNN~\cite{he2017mask} implementation with RoIAlign~\cite{he2017mask}\footnote{\url{https://github.com/roytseng-tw/Detectron.pytorch}} while the region proposal network and the bounding box regression branch are removed because ground-truth object regions are given. 
We use ResNet-50~\cite{he2016deep} as our backbone model. 
Each image is resized with its shorter side 600 pixels.

\subsection{Training} 
We use a stochastic gradient descent with momentum to optimize all the modules. 
The instance-level zero-shot inference and relationship inference modules are trained separately in two stages. 
In the first stage, we train the instance-level zero-shot module on seen categories for 100K iterations. 
The model is fine-tuned from the pretrained ImageNet classification model. %
The learning rate is initialized to 0.005 and reduced by $10\times$ after 60K and 80K iterations. 
After training on the seen categories, we run external algorithms are applied to transfer the knowledge to unseen categories. 
In the second stage, we train the relationship inference module for another 60k iterations with all the other modules fixed.
To facilitate training, we omit unary potentials in Eq.~\eqref{eq:loss} in practice. %
The learning rate is also initialized to 0.005 and reduced by $10\times$ after 20K and 40K iterations. 
For all the modules, the parameter for the weight decay term is set to $0.0001$, and the momentum is 0.9. 
The batch size is set to 8, and the batch normalization layers are fixed during training.

\section{Experiments and results}
\label{sec:exp}

{\renewcommand{\arraystretch}{.8}
\begin{table*}[!tbh]\small
  \caption{Results on Visual Genome dataset. Each group includes two rows. The upper one are baseline methods from zero-shot image classification literature. The lower ones are the results of their models attached with our context-aware inference. HM denotes harmonic mean of the accuracies on $\mathcal{S}$ and $\mathcal{U}$.} \vspace{0.2cm}
  \centering
    \begin{tabular}{l||rr|rr|rr|rr|rr}
    \toprule
          & \multicolumn{2}{c|}{Classic/unseen} & \multicolumn{2}{c|}{Generalized/unseen} & \multicolumn{2}{c|}{Classic/seen} & \multicolumn{2}{c|}{Generalized/seen} & \multicolumn{2}{c}{HM   (Generalized)} \\
 & \multicolumn{1}{c}{per-cls} & \multicolumn{1}{c|}{per-ins} & \multicolumn{1}{c}{per-cls} & \multicolumn{1}{c|}{per-ins} & \multicolumn{1}{c}{per-cls} & \multicolumn{1}{c|}{per-ins} & \multicolumn{1}{c}{per-cls} & \multicolumn{1}{c|}{per-ins} & \multicolumn{1}{c}{per-cls} & \multicolumn{1}{c}{per-ins} \\
    \midrule
    WE    & 18.9  & 25.9  & 3.7   & 3.7   & \textbf{35.6} & \textbf{57.9} & \textbf{33.8} & \textbf{56.1} & 6.7   & 6.9 \\
    WE+Context & \textbf{19.5} & \textbf{28.5} & \textbf{4.1} & \textbf{10.0} & 31.1  & 57.4  & 29.2  & 55.8  & \textbf{7.2} & \textbf{17.0} \\
    \hline
    CONSE & \textbf{19.9} & 27.7  & 0.1   & 0.6   & \textbf{39.8} & 31.7  & \textbf{39.8} & 31.7  & 0.2   & 1.2 \\
    CONSE+Context & 19.6  & \textbf{30.2} & \textbf{5.8} & \textbf{20.7} & 29.6  & \textbf{38.8} & 25.7  & \textbf{35.0} & \textbf{9.5} & \textbf{26.0} \\
    \hline
    GCN   & 19.5  & 28.2  & 11.0  & 18.0  & 39.9  & 31.0  & 31.3  & 22.4  & 16.3  & 20.0 \\
    GCN+Context & \textbf{21.2} & \textbf{33.1} & \textbf{12.7} & \textbf{26.7} & \textbf{41.3} & \textbf{42.4} & \textbf{32.2} & \textbf{35.0} & \textbf{18.2} & \textbf{30.3} \\
    \hline
    SYNC  & 25.8  & 33.6  & 12.4  & 17.0  & 39.9  & 31.0  & 34.2  & 24.4  & 18.2  & 20.0 \\
    SYNC+Context & \textbf{26.8} & \textbf{39.3} & \textbf{13.8} & \textbf{26.5} & \textbf{41.5} & \textbf{39.4} & \textbf{34.5} & \textbf{31.7} & \textbf{19.7} & \textbf{28.9} \\
    \bottomrule
    \end{tabular}%
\label{tab:overall_result}
\end{table*}
}

\subsection{Task}
We mainly evaluate our system on zero shot region classification task. 
We provide ground-truth locations, $\{B_i\}$ for both training and testing.
It enables us to decouple the recognition error from the mistakes from other modules including proposal generation, and diagnose clearly how much context helps zero-shot recognition on object level. 
As a natural extension of our work, we also evaluate on zero-shot detection task. 
In this case, we feed region proposals obtained from Edgeboxes~\cite{zitnick2014edge} instead of ground-truth bounding boxes as input at test time.

\subsection{Dataset}
We evaluate our method on Visual Genome (VG) dataset~\cite{krishna2017visual}, which contains 108K images that have 35 objects and 21 relationships between objects in average.
VG contains two subsets of images, part-1 with around 60K images and part-2 with around 40K images. 
For our experiment, only a subset of categories are considered and the annotated relationships are not directly used.

We use the same seen and unseen category split in \cite{Bansal_2018_ECCV}. 
608 categories are considered for classification. Among these, 478 are seen categories, and 130 are unseen categories. 
The part-1 of VG dataset are used for training, and randomly sampled images from part-2 are used for test. This results in 54,913 training images and 7,788 test images\footnote{The training images still include instances of unseen categories, because pure images with only seen categories are too few. However, we only use annotations of seen categories.}.
The relationship graph in this dataset has 6,396 edges.

\subsection{Metrics and settings}

We employ classification accuracy (AC) for evaluation, where results are aggregated in two ways; ``per-class'' computes the accuracy for each class and then computes the average over all classes while ``per-instance'' is the average accuracy over all regions. 
Intuitively, ``per-class'' metric gives more weight to the instances from rare classes than ``per-instance'' one.

The proposed algorithm is evaluated in both the classic and the generalized zero-shot settings.
The model is only asked to predict among the unseen categories at test time in the classic setting while it needs to consider both seen and unseen categories under generalized setting. 
The generalized setting is more challenging than the classic setting because the model has to distinguish between seen and unseen categories.

\subsection{Baseline methods}
We compare our method with several baselines. 
Note that all baselines treat each object in an image as a separate image thus only utilizing instance-level features for inference.

\vspace{-0.2cm}
\paragraph{Word Embedding (WE)} As described in Section~\ref{sec:instance-level}, a classification is performed by a dot product between a region feature and a weight vector. In this method, weight vector is set to be the GloVe~\cite{pennington2014glove} word embedding of each category. 
Note that the same word embedding is used for the other settings.

\vspace{-0.2cm}
\paragraph{CONSE~\cite{norouzi2013zero}} CONSE first trains classifiers on $\mathcal{S}$ with full supervision.
At test time, each instance in an unseen class is embedded onto the word embedding space by a weighted sum of the seen category embeddings, where the weights are given by the classifier defined on $\mathcal{S}$.
Then the image is predicted to the closest unseen (and seen in the generalized setting) class in the word embedding space.

\paragraph{GCN~\cite{wang2018zero}} 
Similar to CONSE, GCN first trains classifiers on $\mathcal{S}$. Then it learns a GCN model to predict classifier weights for $\mathcal{U}$ from the model for the seen classes. 
The GCN takes the word embeddings of all the seen and unseen categories and the classifier weights of $\mathcal{S}$ as its inputs, and learns the global classifier weights by regression. 
In the end, the predicted classifier weights are used in the inference module for both seen and unseen categories. 
We use a two-layer GCN with LeakyReLU as the activation function. 
Dropout is applied in the intermediate layer and L2 normalization is applied at the output of the network. Following \cite{wang2018zero}, we use WordNet~\cite{miller1995wordnet} to build the graph. 
Each category in VG has its corresponding synset, and is represented as a node in the graph. 
We also add common ancestor nodes of the synsets in VG to connect them in the graph.
In total, 1228 nodes are included in the graph. 

\vspace{-0.2cm}
\paragraph{SYNC~\cite{changpinyo2016synthesized,changpinyo2018classifier}} 
This approach aligns semantic and visual manifolds via use of \textit{phantom} classes. 
The weight of phantom classifier is trained to minimize the distortion error as
\begin{align}
\min_{\mathbf{V}} \|\mathbf{W}_{\mathcal{S}}  - \mathbf{S}_{\mathcal{S}} \mathbf{V} \|,
\end{align}
where $\mathbf{S}_{\mathcal{S}}$ is the semantic similarity matrix between seen categories and phantom classes and $\mathbf{V}$ is the model parameter of the phantom classifier.
The classifier weights for $\mathcal{U}$ is given by a convex combinations of phantom classifier as
\begin{align}
\mathbf{W}_{\mathcal{U}} = \mathbf{S}_{\mathcal{U}} \mathbf{V},
\end{align}
where $\mathbf{S}_{\mathcal{U}}$ is the semantic similarity matrix between unseen categories and phantom classes.

\subsection{Zero-shot recognition results}

Table~\ref{tab:overall_result} presents the performance of our context-aware algorithm based on the four zero-shot recognition baseline methods.
On all backbone baselines, our model improves the accuracy on both unseen categories, both in classic and generalized settings. 
The performances on seen categories are less consistent, which is mainly due to the characteristics of baseline methods, but still better in general.

For the original WE and CONSE methods, we can see that there are huge accuracy gaps between seen and unseen categories, especially under generalized setting. 
This implies that the backbone models are biased towards seen categories significantly. 
Hence, it is natural that our model sacrifices accuracy on $\mathcal{S}$ to improve performance on $\mathcal{U}$.
GCN and SYNC, on the contrary, are more balanced, and our algorithm is able to consistently improve on both seen and unseen categories combined with GCN and SYNC.

The harmonic means of accuracies on seen and unseen categories are consistently higher in our context-aware algorithm than in the baseline methods under generalized setting.
Note that this metric is effective to compare overall performance on both seen and unseen categories as suggested in \cite{xian2018zero}.

\begin{figure}[t]
\begin{minipage}[t]{.45\linewidth}
\setstretch{.7}
\includegraphics[height=2.4cm]{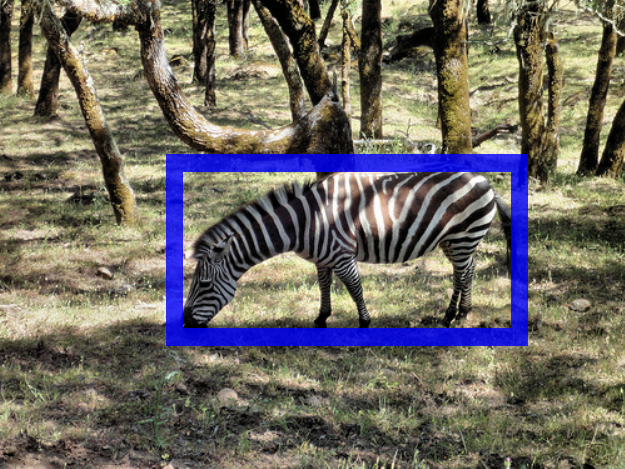}\\
\begin{minipage}[t]{.35\linewidth}
{\footnotesize
animal\\giraffe\\@zebra\\herd\\coat
}
\end{minipage}
\begin{minipage}[t]{.15\linewidth}
{\footnotesize
\textcolor{white}{ }\newline\textcolor{white}{ }\newline$\Rightarrow$
}
\end{minipage}
\begin{minipage}[t]{.4\linewidth}
{\footnotesize
@zebra\\giraffe\\animal\\herd\\coat 
}
\end{minipage}
\end{minipage}%
\hspace{1em}
\begin{minipage}[t]{.45\linewidth}
\setstretch{.7}
\includegraphics[height=2.4cm]{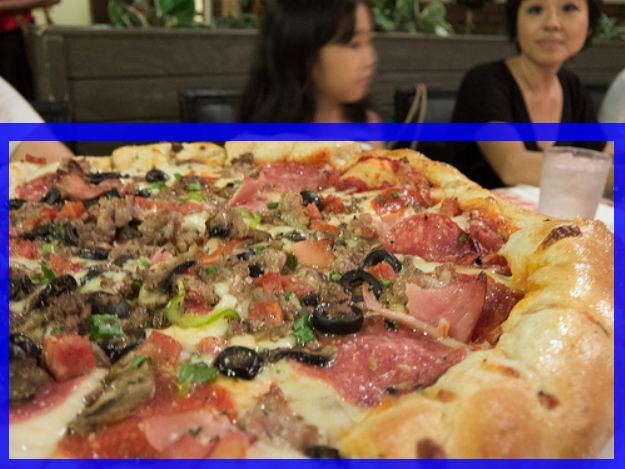}\\
\begin{minipage}[t]{.35\linewidth}
{\footnotesize
pie\\@spatula\\@pizza\\sandwich\\@sugar  
}
\end{minipage}
\begin{minipage}[t]{.15\linewidth}
{\footnotesize
\textcolor{white}{ }\newline\textcolor{white}{ }\newline$\Rightarrow$
}
\end{minipage}
\begin{minipage}[t]{.4\linewidth}
{\footnotesize
@pizza\\sandwich\\pie\\@spatula\\@sugar  
}
\end{minipage}
\end{minipage} \vspace{.5em}
\\
\begin{minipage}[t]{.45\linewidth}
\setstretch{.7}
\includegraphics[height=2.4cm]{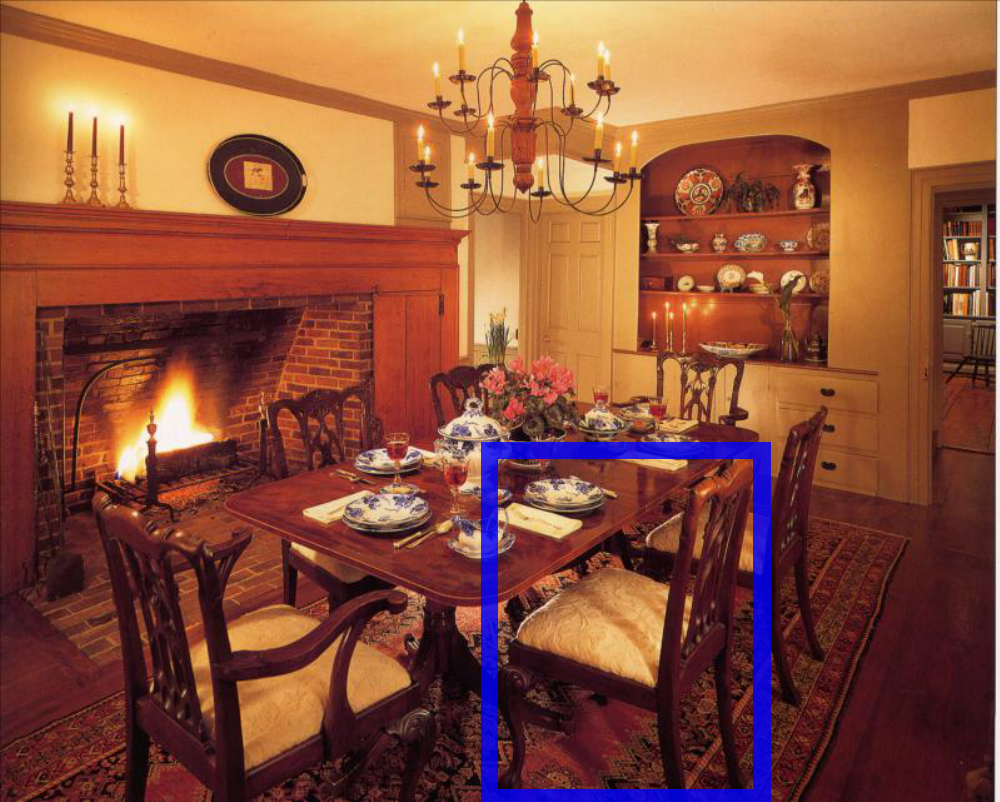}\\
\begin{minipage}[t]{.35\linewidth}
{\footnotesize
furniture\\@chair\\stool\\rug\\@tarpaulin \\
}
\end{minipage}
\begin{minipage}[t]{.15\linewidth}
{\footnotesize
\textcolor{white}{ }\newline\textcolor{white}{ }\newline$\Rightarrow$
}
\end{minipage}
\begin{minipage}[t]{.4\linewidth}
{\footnotesize
@chair\\furniture\\rug\\stool\\@tarpaulin 
}
\end{minipage}
\end{minipage}%
\hspace{1em}
\begin{minipage}[t]{.45\linewidth}
\setstretch{.7}
\includegraphics[height=2.4cm]{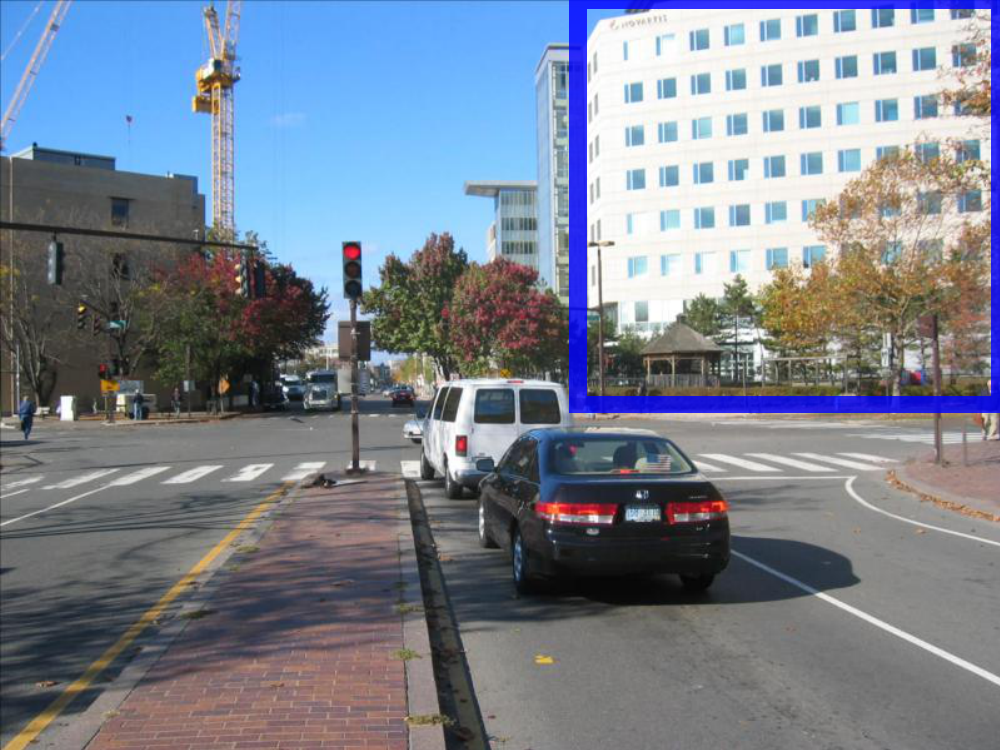}\\
\begin{minipage}[t]{.35\linewidth}
{\footnotesize
skyscraper\\@building\\@house\\sun\\church\\
}
\end{minipage}
\begin{minipage}[t]{.15\linewidth}
{\footnotesize
\textcolor{white}{ }\newline\textcolor{white}{ }\newline$\Rightarrow$
}
\end{minipage}
\begin{minipage}[t]{.4\linewidth}
{\footnotesize
@building\\skyscraper\\@house\\sun\\church
}
\end{minipage}
\end{minipage}
\\
\begin{minipage}[t]{.45\linewidth}
\setstretch{.7}
\includegraphics[height=2.4cm]{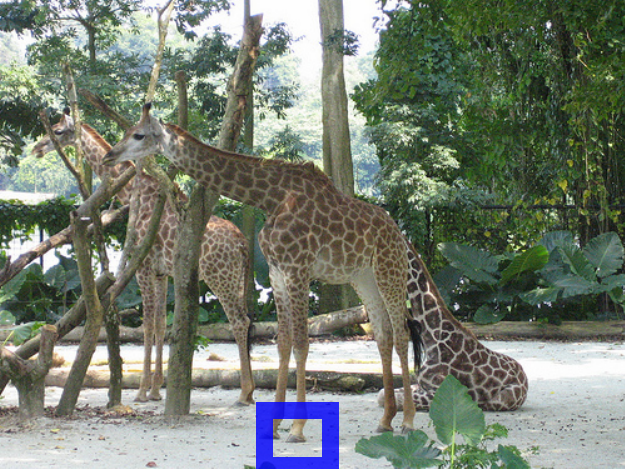}\\
\begin{minipage}[t]{.35\linewidth}
{\footnotesize
@paw\\hoof\\@floor\\pebble\\sand
}
\end{minipage}
\begin{minipage}[t]{.15\linewidth}
{\footnotesize
\textcolor{white}{ }\newline\textcolor{white}{ }\newline$\Rightarrow$
}
\end{minipage}
\begin{minipage}[t]{.4\linewidth}
{\footnotesize
hoof\\@paw\\@floor\\pebble\\sand
}
\end{minipage}
\end{minipage}%
\hspace{1em}
\begin{minipage}[t]{.45\linewidth}
\setstretch{.7}
\includegraphics[height=2.4cm]{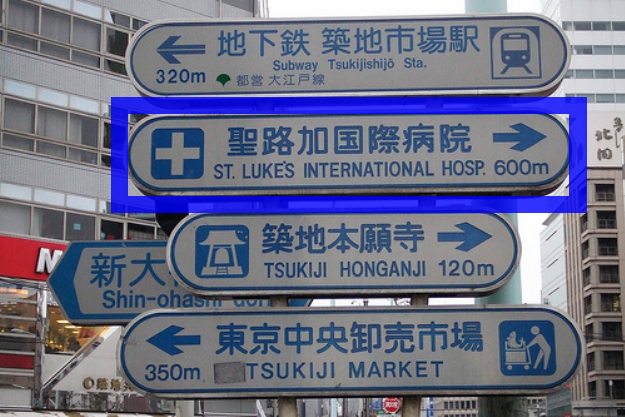}\\
\begin{minipage}[t]{.35\linewidth}
{\footnotesize
billboard\\@poster\\sign\\awning\\@tank
}
\end{minipage}
\begin{minipage}[t]{.15\linewidth}
{\footnotesize
\textcolor{white}{ }\newline\textcolor{white}{ }\newline$\Rightarrow$
}
\end{minipage}
\begin{minipage}[t]{.4\linewidth}
{\footnotesize
sign\\billboard\\@poster\\awning\\@tank
}
\end{minipage}
\end{minipage} \vspace{.5em}
\\ 
\begin{minipage}[t]{.45\linewidth}
\setstretch{.7}
\includegraphics[height=2.4cm]{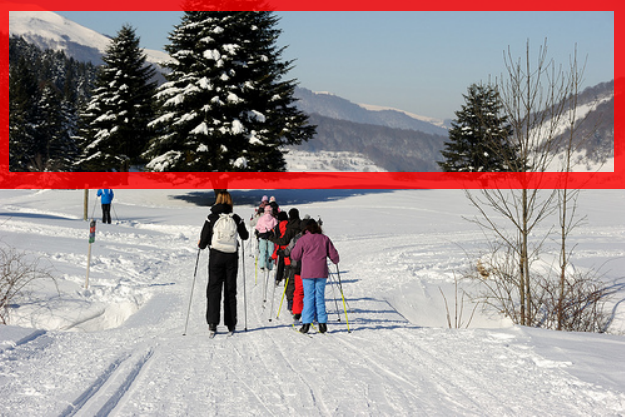}\\
\begin{minipage}[t]{.35\linewidth}
{\footnotesize
@mountain\\ hill\\ @tree\\ lake\\ cliff
}
\end{minipage}
\begin{minipage}[t]{.15\linewidth}
{\footnotesize
\textcolor{white}{ }\newline\textcolor{white}{ }\newline$\Rightarrow$
}
\end{minipage}
\begin{minipage}[t]{.4\linewidth}
{\footnotesize
@tree\\ @mountain\\ hill\\ lake\\ cliff
}
\end{minipage}
\end{minipage}%
\hspace{1em}
\begin{minipage}[t]{.45\linewidth}
\setstretch{.7}
\includegraphics[height=2.4cm]{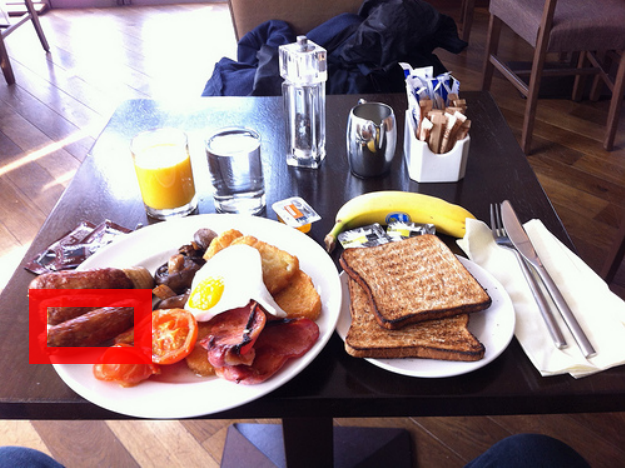}\\
\begin{minipage}[t]{.35\linewidth}
{\footnotesize
@sausage\\ @pepperoni\\ @cake\\ pastry\\ meat
}
\end{minipage}
\begin{minipage}[t]{.15\linewidth}
{\footnotesize
\textcolor{white}{ }\newline\textcolor{white}{ }\newline$\Rightarrow$
}
\end{minipage}
\begin{minipage}[t]{.4\linewidth}
{\footnotesize
@pepperoni\\ meat\\ @cake\\ @sausage\\ pastry
}
\end{minipage}
\end{minipage}
\vspace{1em}
\caption{Examples of top-5 predictions change before (below left) and after (below right) context-aware inference. Blue boxes are examples of correct refinement and red ones denote failure cases. Each unseen category is prefixed with an @ for distinction. }
\label{fig:top-5_refinement}
\vspace{-1em}
\end{figure}

\begin{figure*}[!th]
\hspace{1.0em}
\begin{minipage}[t]{.23\linewidth}
\setstretch{.7}
\includegraphics[height=2.6cm]{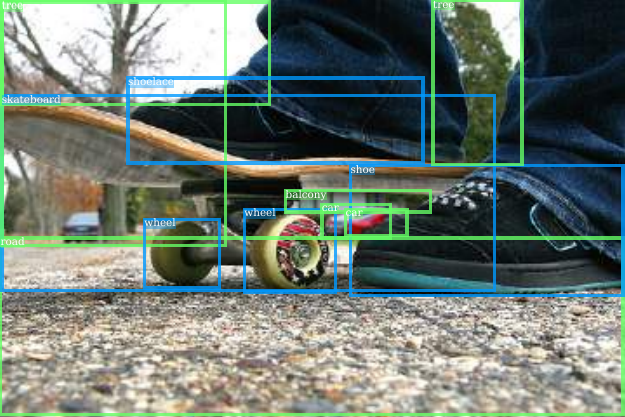}\\
\end{minipage}%
\hspace{.5em}
\begin{minipage}[t]{.23\linewidth}
\setstretch{.7}
\includegraphics[height=2.6cm]{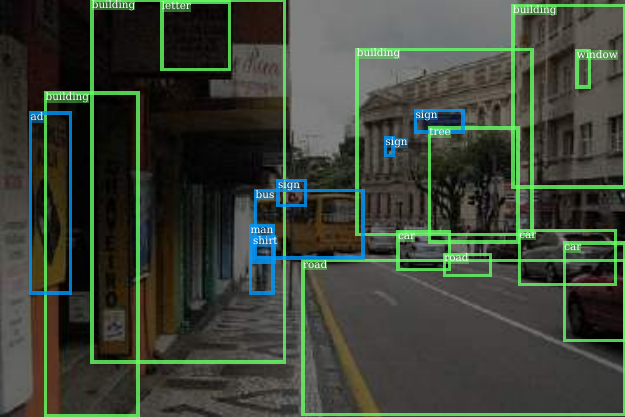}\\
\end{minipage}
\hspace{.5em}
\begin{minipage}[t]{.23\linewidth}
\setstretch{.7}
\includegraphics[height=2.6cm]{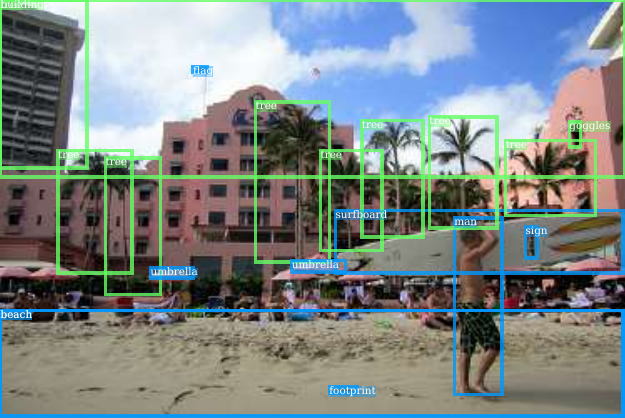}\\
\end{minipage}
\hspace{.5em}
\begin{minipage}[t]{.23\linewidth}
\setstretch{.7}
\includegraphics[height=2.6cm]{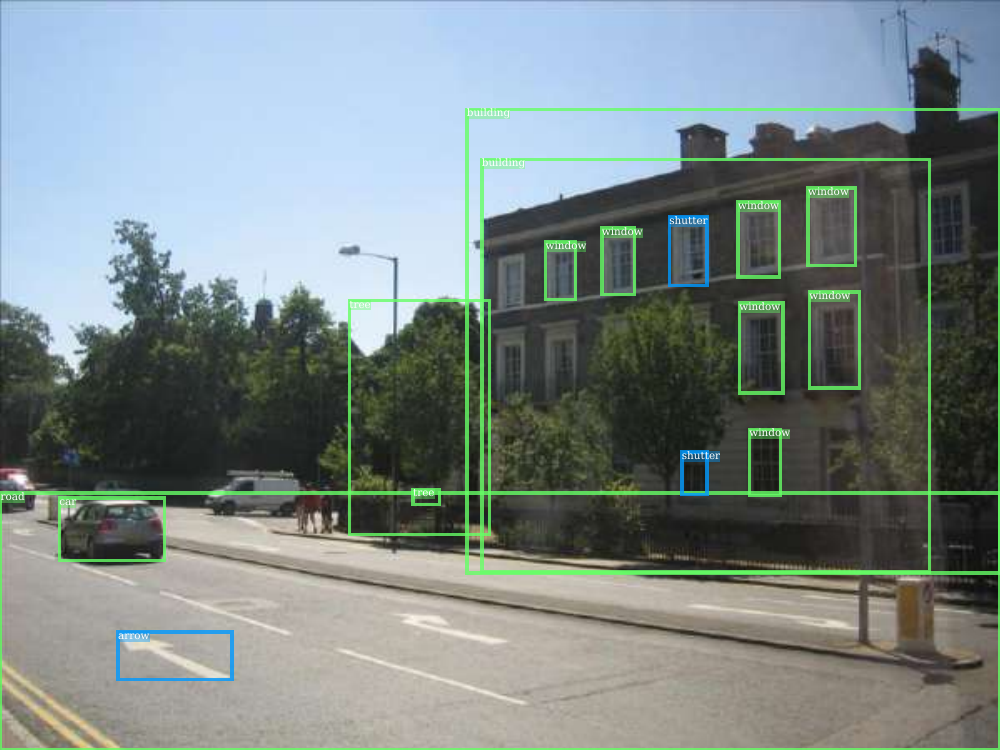}\\
\end{minipage}
\caption{More qualitative results for zero shot region classification.
The blue and green bounding boxes corresponds to objects of seen and unseen categories, respectively.
}
\label{fig:sample_recognition}
\end{figure*}

\begin{table*}[t]%
  \centering
  \caption{Results of different inputs to relationship inference module. *+G is the model with only geometry information. *+GA is the model with both geometry and appearance feature.}\label{tab:geoablation}
    \vspace{.5em}
   \scalebox{0.9}{
   \begin{tabular}{l||rr|rr|rr|rr}
   \toprule
          & \multicolumn{2}{c|}{Classic/unseen} & \multicolumn{2}{c|}{Generalized/unseen} & \multicolumn{2}{c|}{Classic/seen} & \multicolumn{2}{c}{Generalized/seen} \\
& \multicolumn{1}{r}{per-cls} & \multicolumn{1}{r|}{per-ins} & \multicolumn{1}{r}{per-cls} & \multicolumn{1}{r|}{per-ins} & \multicolumn{1}{r}{per-cls} & \multicolumn{1}{r|}{per-ins} & \multicolumn{1}{r}{per-cls} & \multicolumn{1}{r}{per-ins} \\
\hline
    GCN   & 19.5  & 28.2  & 11.0  & 18.0  & 39.9  & 31.0  & 31.3  & 22.4 \\
    GCN+G & \textbf{21.2} & \textbf{33.1} & \textbf{12.7} & \textbf{26.7} & \textbf{41.3} & 42.4  & 32.2  & 35.0 \\
    GCN+GA & 20.4  & 26.5  & 9.2   & 15.3 & 40.9 & \textbf{44.8} & \textbf{34.7} & \textbf{40.9} \\
    \hline
    SYNC  & 25.8  & 33.6  & 12.4  & 17.0  & 39.9  & 31.0  & 34.2  & 24.4 \\
    SYNC+G & \textbf{26.8} & \textbf{39.3} & \textbf{13.8} & \textbf{26.5} & 41.5  & 39.4  & 34.5  & 31.7 \\
    SYNC+GA & 26.6  & 33.6  & 11.3  & 16.4  & \textbf{41.6} & \textbf{42.8} & \textbf{36.5} & \textbf{38.5} \\
    \bottomrule
    \end{tabular}%
    }
\end{table*}

{\renewcommand{\arraystretch}{.7}
\begin{table}[!t]\small
  \centering
    \caption{Per-instance top-$K$ accuracy on unseen categories.}
    \vspace{.5em}
    \begin{tabular}{l||cc|cc}
    \toprule
          & \multicolumn{2}{c|}{Generalized} & \multicolumn{2}{c}{Classic} \\
& \multicolumn{1}{c}{top-1} & \multicolumn{1}{c|}{top-5} & \multicolumn{1}{c}{top-1} & \multicolumn{1}{c}{top-5} \\
    \midrule
    WE+Ctx & {\color{white}0}3.7 $\rightarrow$ 10.0 & 26.6  & 25.9 $\rightarrow$ 28.5 & 57.5 \\
    CONSE+Ctx & {\color{white}0}0.6 $\rightarrow$ 20.7 & 29.4  & 27.7 $\rightarrow$ 30.2 & 56.1 \\
    GCN+Ctx & 18.0 $\rightarrow$ 26.7 & 38.3  & 28.2 $\rightarrow$ 33.1 & 51.6 \\
    SYNC+Ctx & 17.0 $\rightarrow$ 26.5 & 49.4  & 33.6 $\rightarrow$ 39.3 & 68.9 \\
    \bottomrule
    \end{tabular}
    \vspace{.5em}
\label{tab:top-1top-5}
\end{table}
}

\vspace{-0.2cm}
\paragraph{Top-$K$ refinement} 
As we mentioned in Section~\ref{sec:context-aware}, our pruning method makes the context-aware inference a top-$k$ class reranking. 
We conduct current experiment with $K=5$, results with other options of $K$ can be seen in \supporapp. 
In Table \ref{tab:top-1top-5}, we show ``per-instance'' top-1 accuracy versus top-5 accuracy of different algorithms on unseen categories. 
The top-5 accuracies are not changed since we only rerank the top-5 classes, and the top-1 accuracy we can achieve are upper bounded by the corresponding top-5 accuracy. After applying context-aware inference, the top-1 accuracies increase. 
Notably, the baseline model of CONSE has near 0 accuracy under generalized setting because it biases towards seen categories severely. 
However, its top-5 accuracy is reasonable. 
Our method is able to reevaluate top-5 predictions with the help of relation knowledge and increase the top-1 accuracy significantly.

\vspace{-0.2cm}
\paragraph{Qualitative results} 
Figure \ref{fig:top-5_refinement} shows qualitative results from the context-aware inference. 
Our context-aware model adjusts the class probabilities based on the object context.
For example, zebra is promoted in the first image because the bands on its body while sausage helps recognize the pizza in the second image. 
Different patterns can be found for label refinement: general to specific (furniture to chair, air craft to airplane, animal to zebra), specific to general (skyscraper to building), and corrected to similar objects (pie to pizza, paw to hoof). 
Figure~\ref{fig:sample_recognition} shows more qualitative results of region classification after applying context-aware inference.

\vspace{-0.2cm}
\paragraph{Input choices for relationship inference} Our relationship inference module only takes geometry information as input to avoid overfitting to seen categories. One alternative we tried is combining it with region appearance feature. We project region features $f_i$ and $f_j$ into lower dimension and concatenate it with $\mathcal E(g_{ij})$ to produce relation potentials. We report the results in Table \ref{tab:geoablation}. The appearance augmented relationship inference module is named as +GA in the table. It's shown that +GA biases towards seen categories, and hurts performance on unseen categories. +GA on generalized setting on unseen categories is even worse than the baselines.

\paragraph{Results by varying the size of $\mathcal{S}$}
We generate several subsets of $\mathcal{S}$ by subsampling with the ratios of $1/2$ and $1/5$, while the unseen category set remains the same.
Table~\ref{table:seen_unseen} shows that our context-aware method consistently benefits in the zero-shot recognition in this ablation study.
\begin{table}[t]
  \centering
  \caption{Results by varying the size of $\mathcal{S}$ in terms of per-cls accuracy on $\mathcal{U}$ in classic setting.} 
  \vspace{0.1cm}
  \label{table:seen_unseen}
  \scalebox{0.9}{
  \setlength\tabcolsep{3pt}
    \begin{tabular}{c||ccc|ccc}
    \toprule
    {\# of seens} & \multicolumn{1}{c}{GCN} & \multicolumn{1}{c}{GCN+Ctx} & \multicolumn{1}{c|}{$\Delta$} & \multicolumn{1}{c}{SYNC} & \multicolumn{1}{c}{SYNC+Ctx} & \multicolumn{1}{c}{$\Delta$} \\
    \midrule
    ~~95 (20\%)   & ~~7.2   & ~~7.7   & 0.5   & 10.6  & 10.9  & 0.3 \\
    239 (50\%)   & 13.8  & 14.2  & 0.4   & 19.5  & 19.5  & 0.0 \\
    478 (100\%)  & 19.5  & 21.2  & 1.7   & 25.8  & 26.8 & 1.0 \\
    \bottomrule
    \end{tabular}%
    }
  \vspace{-.5em}
\end{table}

\subsection{Zero-shot detection results}
We extend our region classification model for detection task by adding a background detector. 
We set the classifier weight of background class to be normalized average classifier weights: 
$$ \mathbf{W}_{\text{bg}} = \frac{\sum_{c\in \mathcal C} \mathbf{W}_c}{\|\sum_{c\in \mathcal C} \mathbf{W}_c||^2},$$ 
where each row of $\mathbf{W}$ needs to be normalized in advance. 
Furthermore, given thousands of region proposals, we only consider the top 100 boxes with highest class scores given by instance-level module for context-aware inference.

Following \cite{Bansal_2018_ECCV}, EdgeBoxes proposals are extracted for test images, where only proposals with scores higher than 0.07 are selected. 
After detection, non-maximum suppression is applied with IOU threshold 0.4. 
Due to incomplete annotations in VG, we report Recall@100 scores with IOU threshold 
0.4/0.5.
Table \ref{tab:zsd} presents instance-level zero-shot performance of GCN and SYNC models, where our method shows improved accuracy on unseen categories and higher overall recalls given by harmonic means.
Note that our results on the generalized zero-shot setting already outperforms the results on the classic setting reported in \cite{Bansal_2018_ECCV}.

\begin{table}[t]%
  \centering
    \caption{Generalized zero-shot detection results.
    Recall@100 with IOU threshold 0.4/0.5 is reported.} \vspace{.5em}
    \scalebox{0.88}{
      \begin{tabular}{l||cc|cc|cc}
      \toprule
          & \multicolumn{2}{c|}{Unseen} & \multicolumn{2}{c|}{Seen} & \multicolumn{2}{c}{Harmonic mean} \\
          & {0.4} & {0.5}  & {0.4} & {0.5}  & {0.4} & {0.5}  \\
    \midrule
    GCN   & ~~8.5   & 6.2    & \textbf{23.1} & \textbf{17.8}  & 12.4  & ~~9.2   \\
    GCN+Context & \textbf{~~9.7} & \textbf{6.9}  & 22.3  & 16.0    & \textbf{13.5} & \textbf{~~9.6} \\
    \hline
    SYNC  & 11.1  & 8.2   & \textbf{24.2} & \textbf{18.8} & 15.2  & 11.4   \\
    SYNC+Context & \textbf{12.0} & \textbf{8.6} & 23.1  & 17.4   & \textbf{15.8} & \textbf{11.5}\\
    \bottomrule
    \end{tabular}%
    }
\label{tab:zsd}
\end{table}

%% file: sections/conclusion.tex
\section{Conclusions}\label{sec:conclusion}
We presented a novel setting for zero-shot object recognition, where high-level visual context information is employed for inference. 
Under this setting, we proposed a novel algorithm to incorporate both instance-level and object relationship knowledge in a principled way. 
Experimental results show that our context-aware approach boosts the performance significantly compared to the models with only instance-level information. 
We believe that this new problem setting and the proposed algorithm facilitate more interesting research for zero-shot or few-shot learning.

%% file: sections/appendix.tex
We first analyze the effect of $k$, which is the number of top categories picked for efficient inference in our CRF module (Section~3.2.3). 
In the second part, we describe how hyperparameter $\gamma$ is chosen. 
In the third part, we analyze the accuracy improvement pattern of our model for different object categories. In the fourth part, we visualize the learned relation potentials $\ell_r^{ij}$ and pairwise potentials $\theta_{ij}$.

\section{Effect of $k$ in pruning}
For efficiency, we use only top-$k$ categories for inference in the mean field algorithm described in Section (3.2.3). 
Here we show that the choice of $k$ only affects the accuracy of our algorithm marginally. 
Table \ref{tab:topk} shows the result of four baseline models with different choices of $k$. 
We notice higher $k$ leads to slightly worse performance on seen categories, but improves on unseen categories in general.
\begin{table*}[!tbh]\small
  \centering
    \begin{tabular}{l|rr|rr|rr|rr|rr}
    \multicolumn{1}{r}{} & \multicolumn{2}{c|}{Classic/unseen} & \multicolumn{2}{c|}{Generalized/unseen} & \multicolumn{2}{c|}{Classic/seen} & \multicolumn{2}{c|}{Generalized/seen} & \multicolumn{2}{c}{HM(Generalized)} \\
\cmidrule{2-11}    \multicolumn{1}{r}{} & \multicolumn{1}{c}{per-cls} & \multicolumn{1}{c|}{per-ins} & \multicolumn{1}{c}{per-cls} & \multicolumn{1}{c|}{per-ins} & \multicolumn{1}{c}{per-cls} & \multicolumn{1}{c|}{per-ins} & \multicolumn{1}{c}{per-cls} & \multicolumn{1}{c|}{per-ins} & \multicolumn{1}{c}{per-cls} & \multicolumn{1}{c}{per-ins} \\
\midrule
    WE+Context(5) & 19.5  & 28.5  & 4.1   & 10.0  & 31.1  & 57.4  & 29.2  & 55.8  & 7.2   & 17.0 \\
    WE+Context(10) & 20.1  & 33.1  & 4.1   & 11.3  & 30.0  & 56.7  & 28.1  & 55.0  & 7.2   & 18.7 \\
    WE+Context(20) & 20.5  & 36.0  & 4.0   & 11.6  & 29.5  & 56.2  & 27.6  & 54.3  & 7.0   & 19.1 \\
    \midrule
    CONSE+Context(5) & 19.6  & 30.2  & 5.8   & 20.7  & 29.6  & 38.8  & 25.7  & 35.0  & 9.5   & 26.0 \\
    CONSE+Context(10) & 18.6  & 32.7  & 6.0   & 23.3  & 23.9  & 36.4  & 19.5  & 31.2  & 9.2   & 26.7 \\
    CONSE+Context(20) & 16.6  & 33.0  & 5.3   & 22.1  & 18.3  & 32.5  & 14.1  & 26.4  & 7.7   & 24.1 \\
    \midrule
    GCN+Context(5) & 21.2  & 33.1  & 12.7  & 26.7  & 41.3  & 42.4  & 32.2  & 35.0  & 18.2  & 30.3 \\
    GCN+Context(10) & 21.8  & 35.6  & 12.7  & 27.9  & 40.3  & 45.3  & 30.9  & 36.4  & 18.0  & 31.6 \\
    GCN+Context(20) & 21.4  & 36.7  & 12.0  & 28.1  & 39.3  & 45.7  & 30.0  & 36.3  & 17.1  & 31.7 \\
    \midrule
    SYNC+Context(5) & 26.8  & 39.3  & 13.8  & 26.5  & 41.5  & 39.4  & 34.5  & 31.7  & 19.7  & 28.9 \\
    SYNC+Context(10) & 27.2  & 41.6  & 13.8  & 27.1  & 41.3  & 41.2  & 34.4  & 32.4  & 19.7  & 29.5 \\
    SYNC+Context(20) & 27.2  & 42.2  & 13.9  & 27.1  & 41.2  & 41.6  & 34.2  & 32.4  & 19.7  & 29.5 \\
    \bottomrule
    \end{tabular}%
    \vspace{.5em}
  \caption{Performance of our model with different top-$k$ settings for CRF inference. The number in the parentheses is the parameter $k$ for each setting, respectively.}\label{tab:topk}
\end{table*}

\vspace{-0.2cm}
\paragraph{Computation cost} Relative to runtime without context inference, top-100 and top-5 pruning increase the runtime by $\small \sim$58\% and $\small \sim$18\% respectively. If without pruning, out-of-memory error will be raised.

\section{Choice of $\gamma$}
We split the original seen label set into dev\_seen and dev\_unseen evenly; $\gamma$ is chosen to be the best performing one on dev set.
For WE, CONSE, GCN, $\gamma$ is $1$, and for SYNC, $\gamma$ is set to be $0.5$.

\section{Accuracy improvement for different classes}
In the experiment section, we see that our algorithm improves more on `per-instance' than `per-class' metric. In order to investigate this outcome, we analyze the correlation between accuracy improvement of individual categories and two factors: degree of the category in the relation graph and frequency of the category.
The model we use in this section is GCN+Context.

\begin{figure}[t]
\centering
\includegraphics[width=0.95\linewidth]{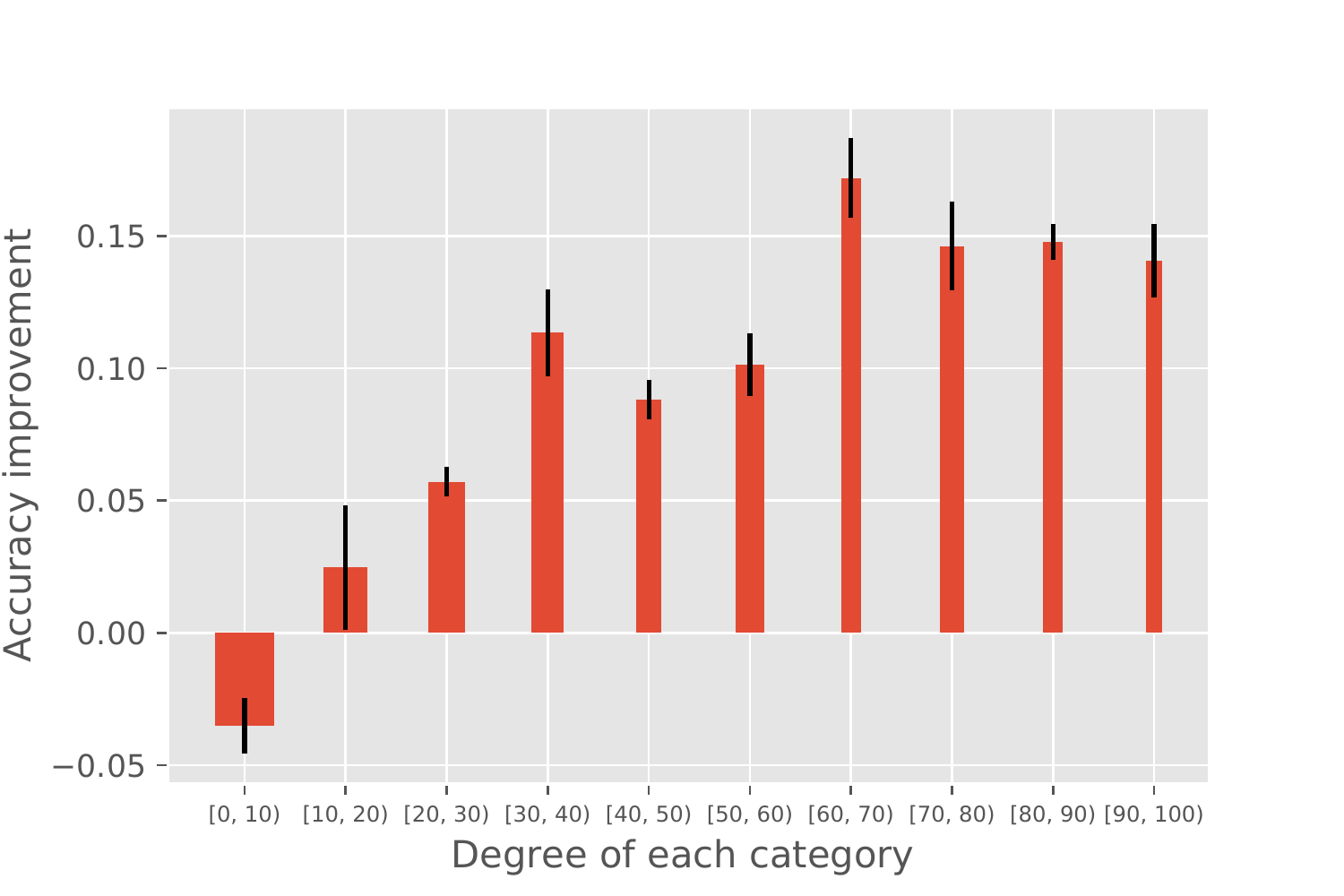}
\caption{Correlation between accuracy improvement of individual categories and degree of the category in the relation graph. The width of each bar is proportional to logarithm of number of categories in the bin. $x$-axis denotes degree of a category in the graph.}
\label{figs:graph_degree}
\end{figure}

In Fig.~\ref{figs:graph_degree}, we analyze the correlation between accuracy improvement of individual categories and degree of the category in the relation graph. \ljcom{which model is it}\rtcom{it's GCN+Context} Here the degree of one category is equal to the number of relationships where the category is either subject or object. $x$-axis is the degree of each category in the graph, while $y$-axis is the relative accuracy improvement compared to the baseline model. It is shown that for classes with high degrees, the accuracies are mostly improved; for classes with low degrees, the accuracies actually drop a little bit on average. In general, since most categories are improved, the overall `per-class' accuracy is improved.

\begin{figure}[t]
\centering
\includegraphics[width=\linewidth]{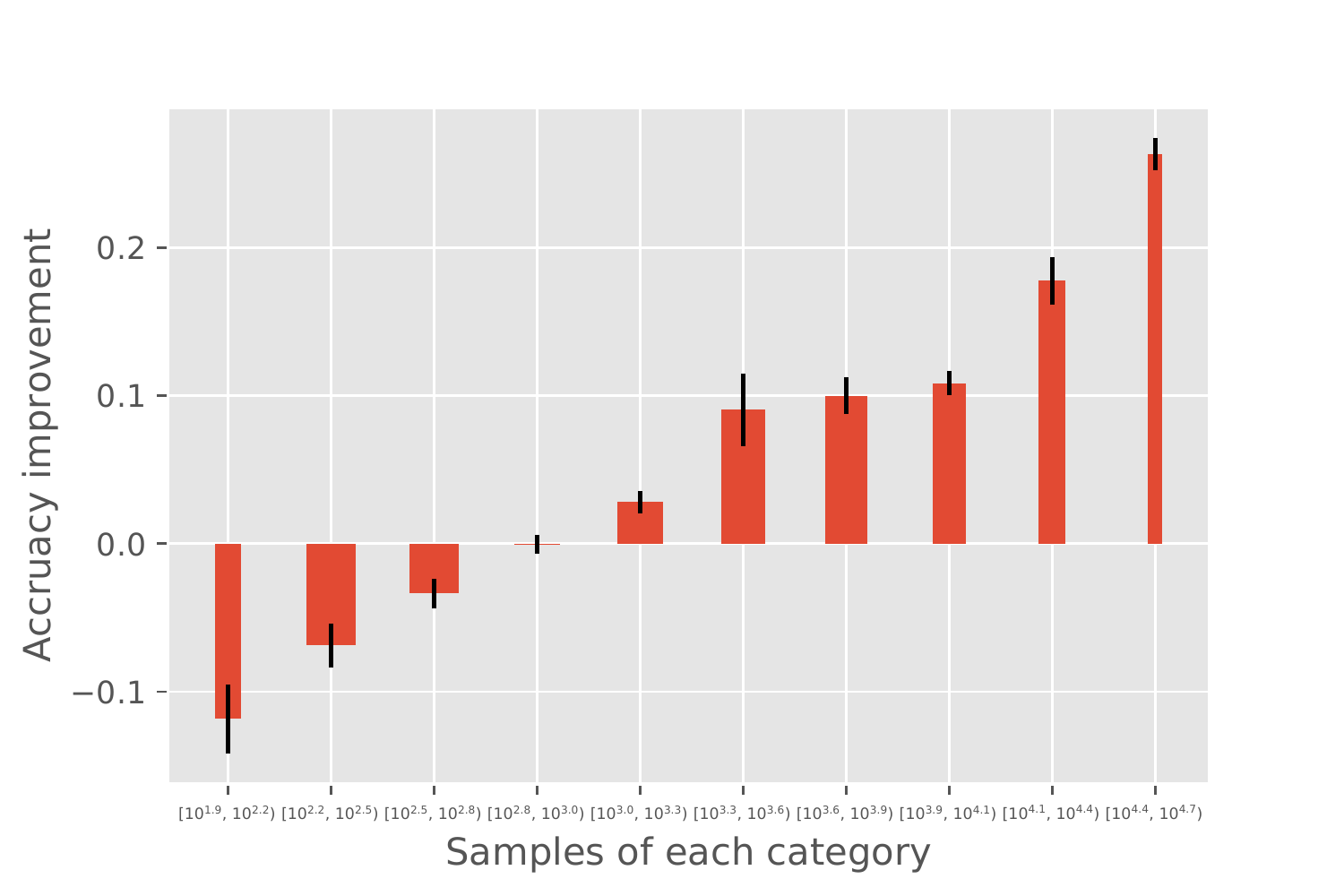}
\caption{Correlation between accuracy improvement and category frequencies. The width of each bar is proportional to logarithm of number of categories in the bin. $x$-axis represents number of samples of a category in the test set.}
\label{figs:frequency}
\end{figure}

In Fig.~\ref{figs:frequency}, we analyze the correlation between accuracy improvement and category frequencies. $x$-axis is the number of samples for each category in the whole test set (number in training set is not available for unseen categories). We can see that categories with more occurrences in the test set have larger improvement. This is why more gains are obtained on `per-instance' compared to `per-class' metric. Generally, categories with more samples have more relation/interactions with other object categories thus providing more cues to be inferred from a context. 

\section{Visualization of relation potentials $\ell_r^{ij}$ and pairwise potentials $\theta_{ij}$}
We provide visualizations of learned relation knowledge of our algorithm. The model we use in this section is GCN+Context.

\begin{figure}[t]
\centering
\begin{minipage}[t]{.99\linewidth}
\begin{minipage}[t]{.49\linewidth}
\includegraphics[height=2.7cm]{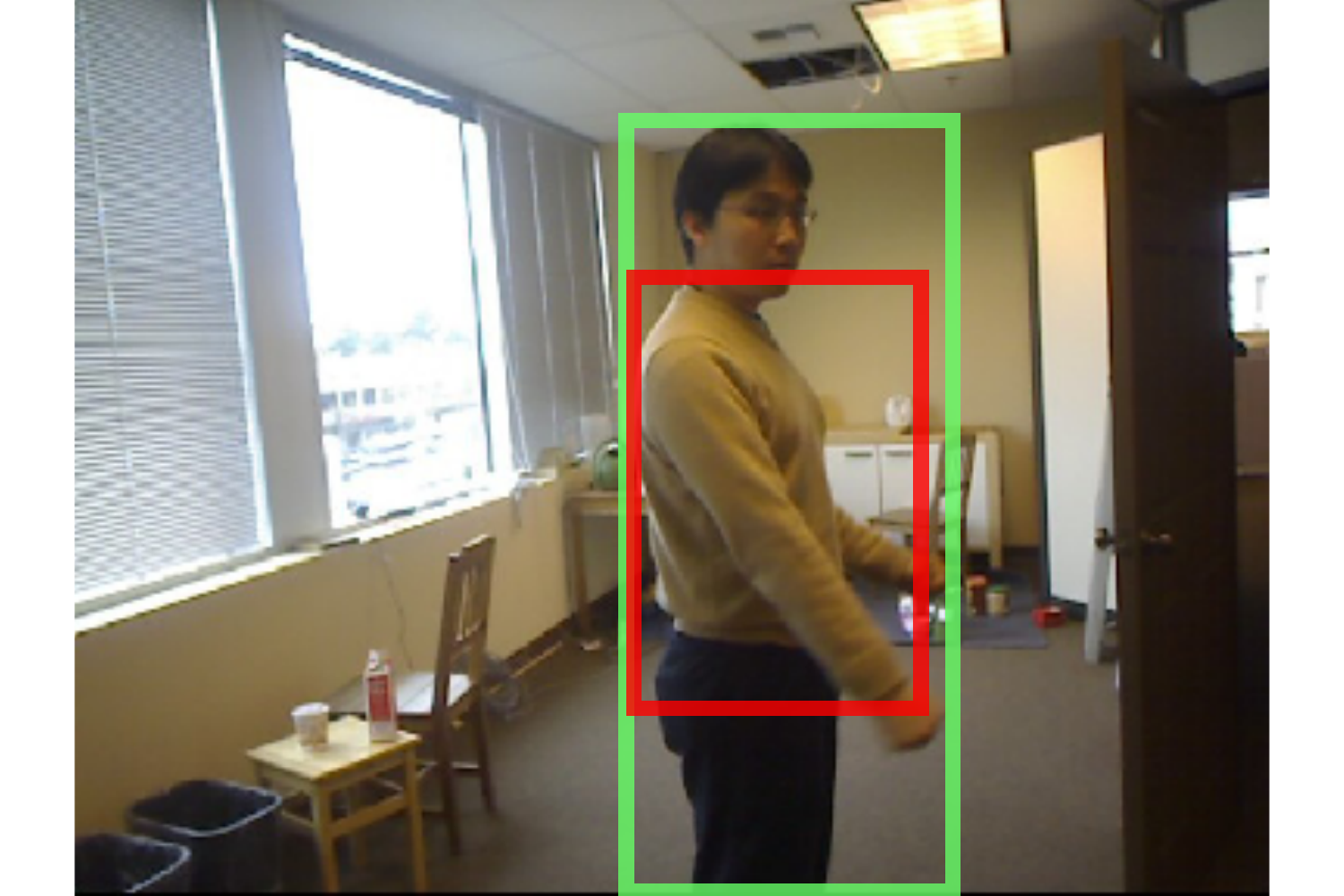}
\end{minipage}
\begin{minipage}[t]{.49\linewidth}
\includegraphics[height=2.7cm]{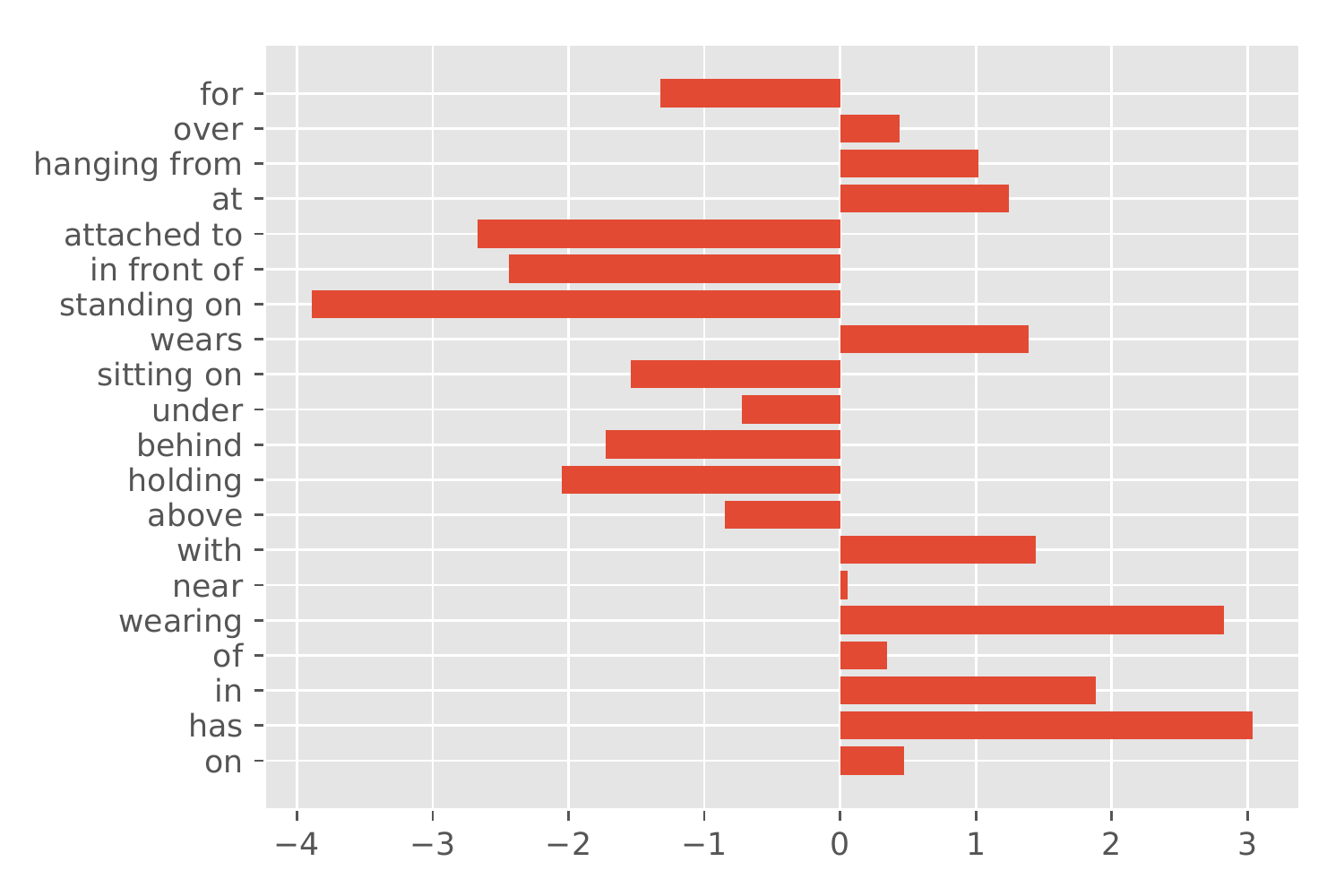}
\end{minipage}
\end{minipage}

\begin{minipage}[t]{.99\linewidth}
\begin{minipage}[t]{.49\linewidth}
\includegraphics[height=2.7cm]{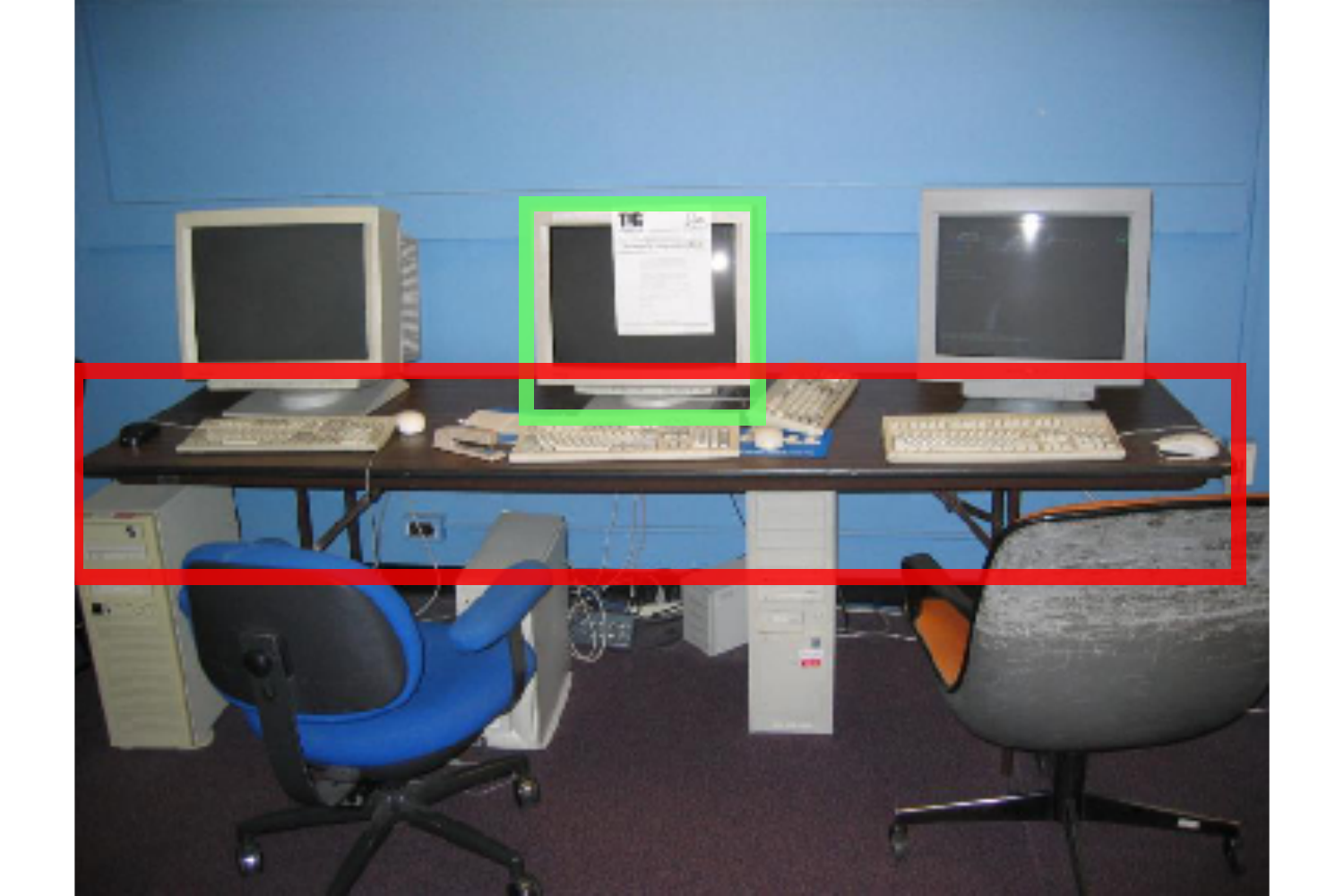}
\end{minipage}
\begin{minipage}[t]{.49\linewidth}
\includegraphics[height=2.7cm]{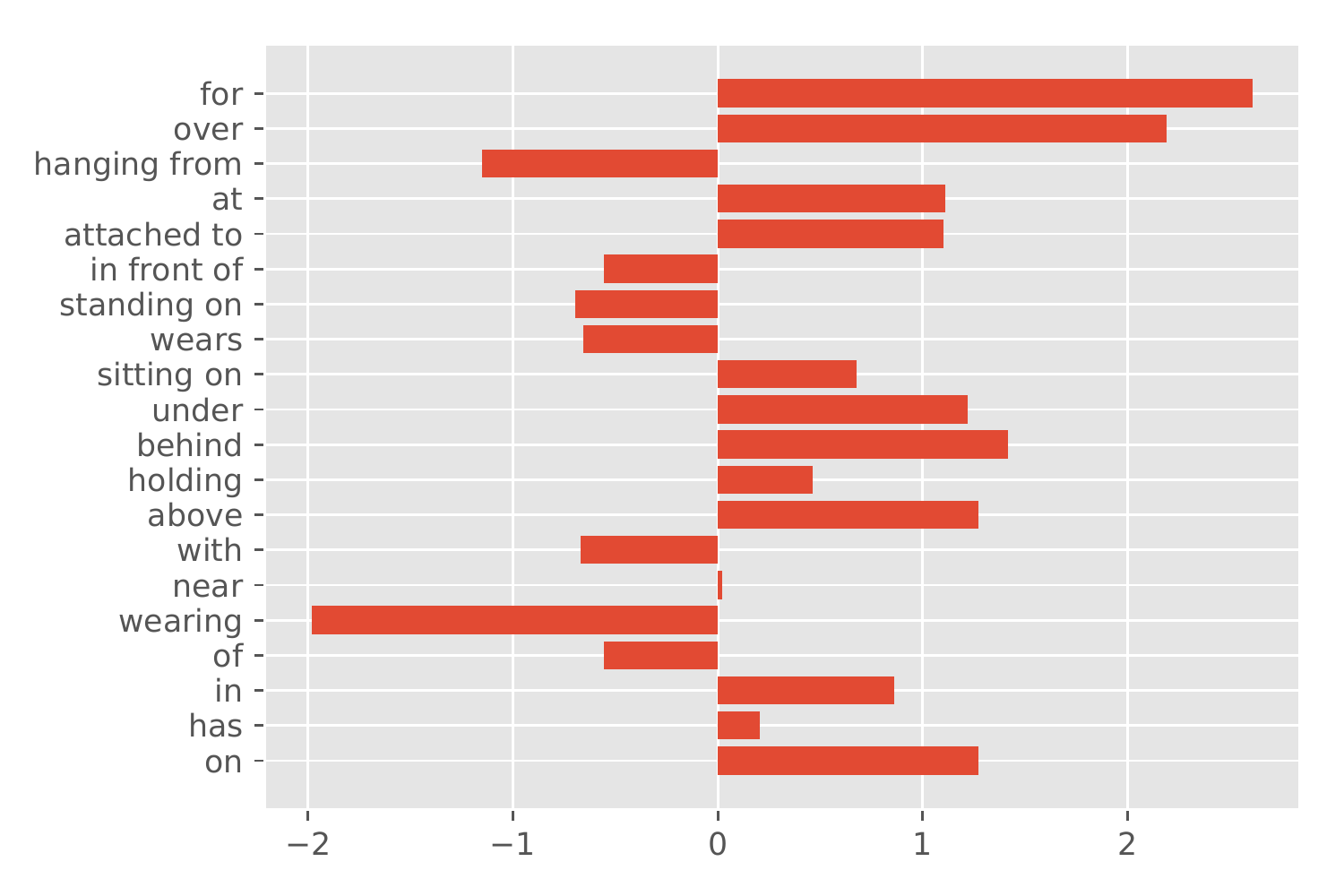}
\end{minipage}
\end{minipage}

\begin{minipage}[t]{.99\linewidth}
\begin{minipage}[t]{.49\linewidth}
\includegraphics[height=2.7cm]{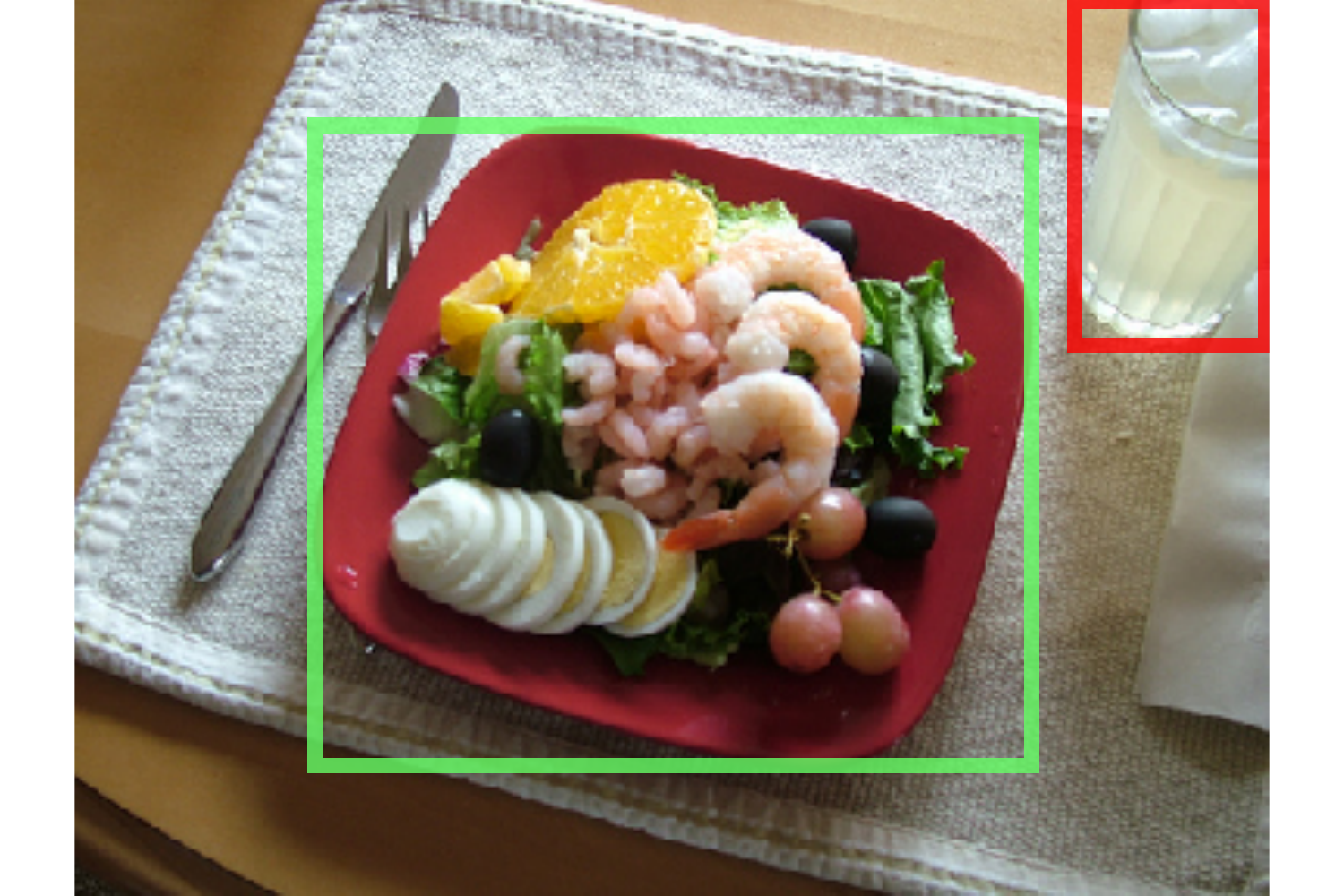}
\end{minipage}
\begin{minipage}[t]{.49\linewidth}
\includegraphics[height=2.7cm]{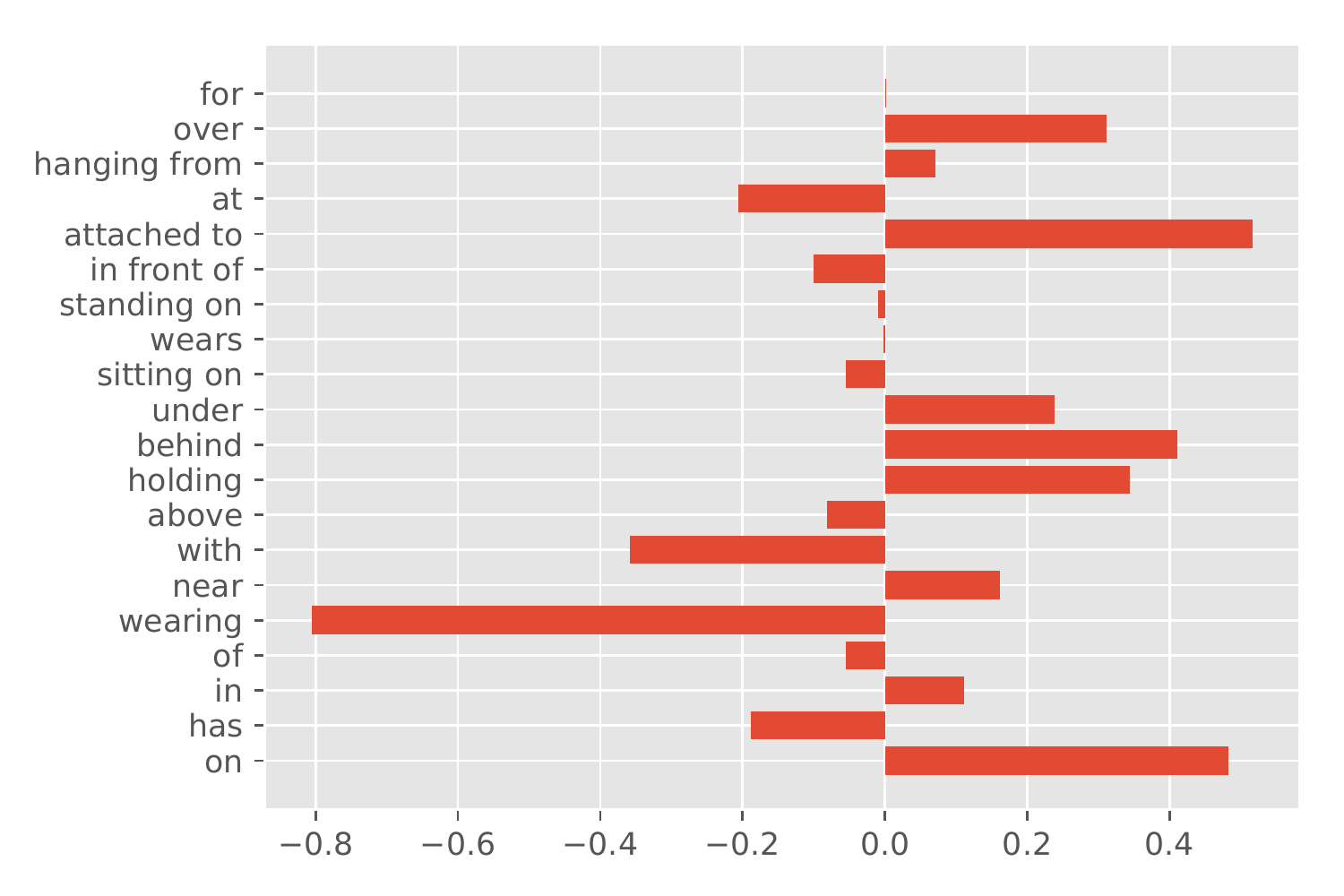}
\end{minipage}
\end{minipage}

\begin{minipage}[t]{.99\linewidth}
\begin{minipage}[t]{.49\linewidth}
\includegraphics[height=2.7cm]{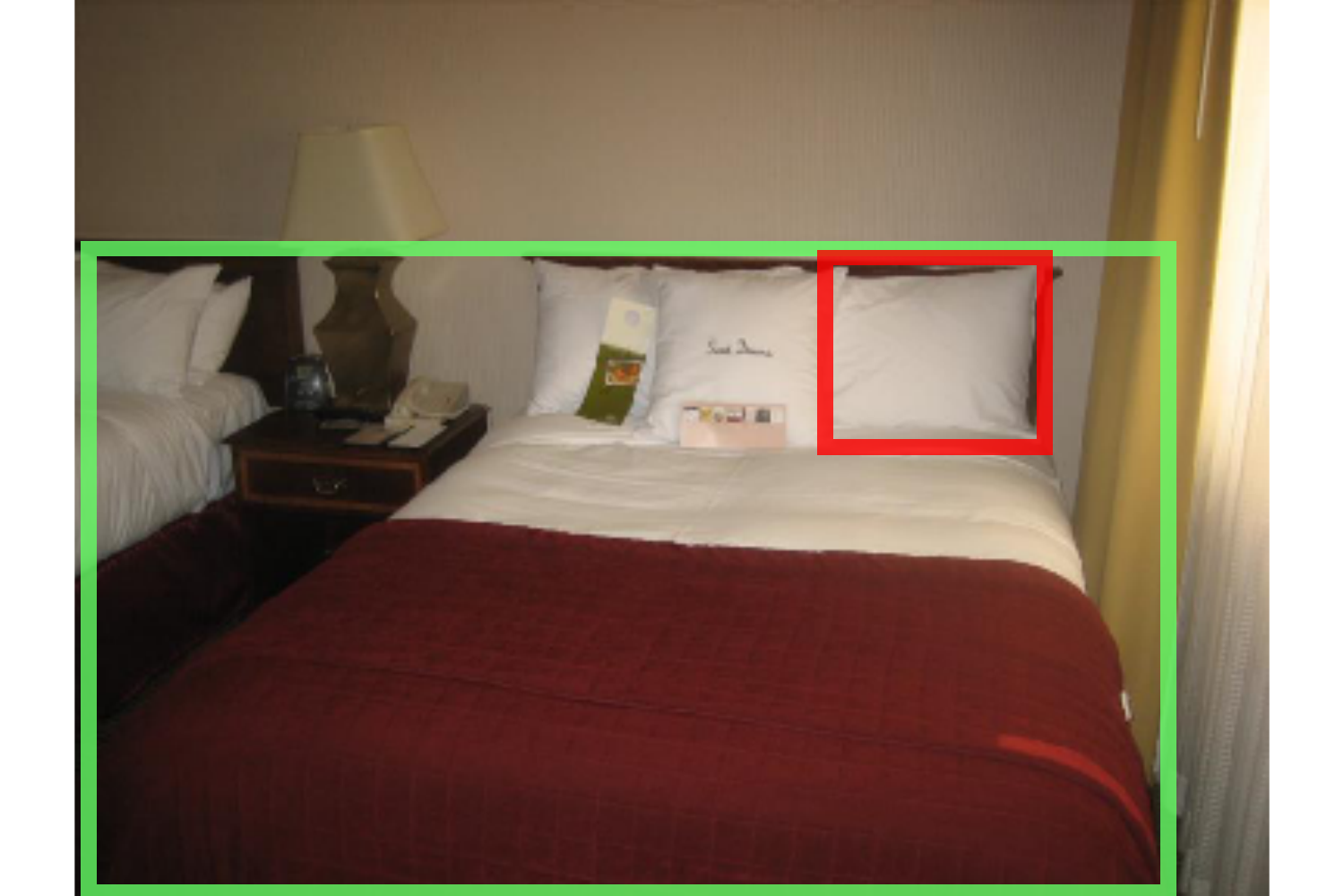}
\end{minipage}
\begin{minipage}[t]{.49\linewidth}
\includegraphics[height=2.7cm]{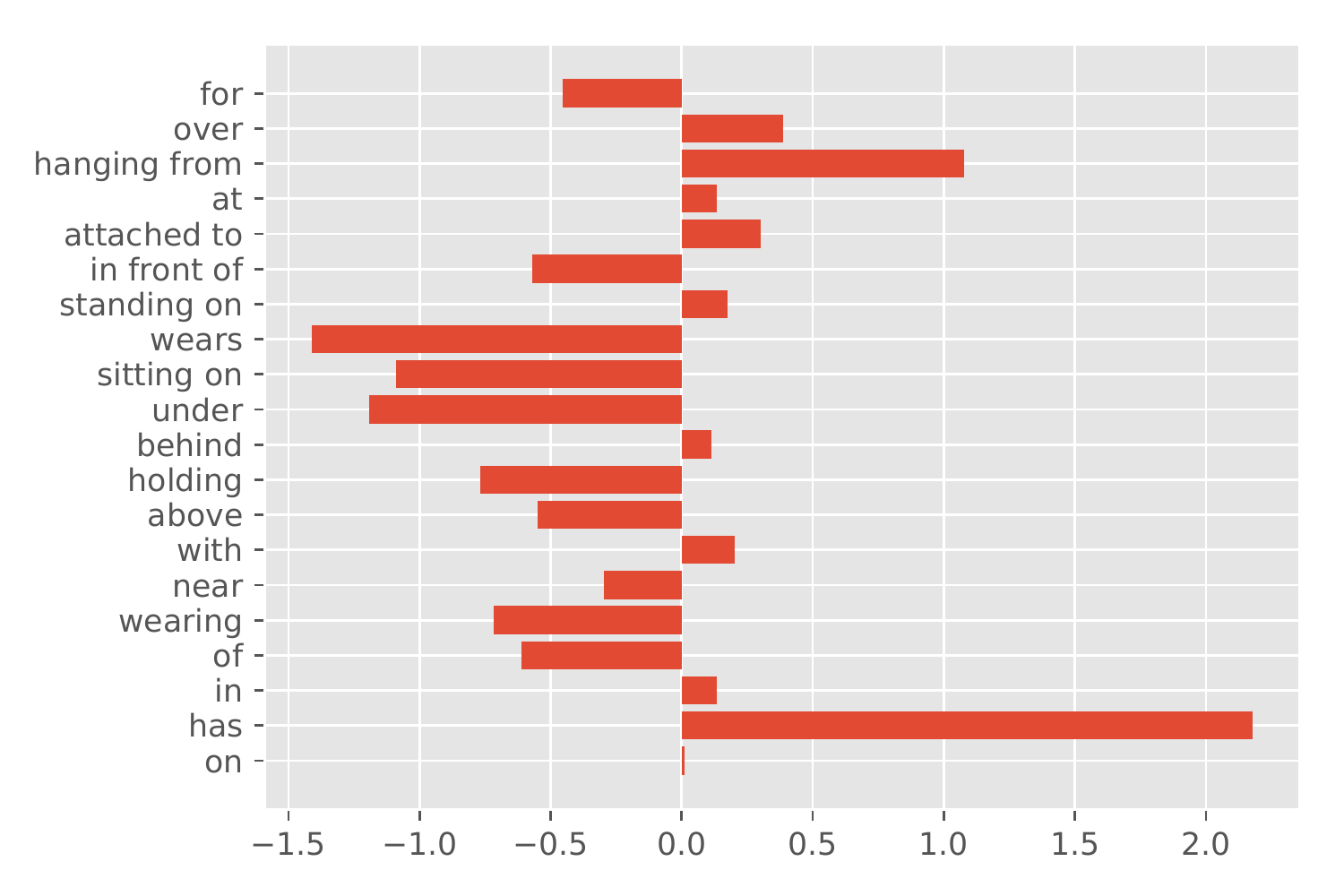}
\end{minipage}
\end{minipage}
\vspace{0.2cm}
\caption{Visualization of relation potentials. For a pair of objects, green box denotes subject and red one denotes object. The values of potential $\ell_r^{i,j}$ are shown on the right side of each image, respectively.}
\label{figs:pairs}
\end{figure}

In Fig. \ref{figs:pairs}, we illustrate the relation potentials $\ell_r^{ij}$ given the location of subject and object. Our model is able to learn pairwise `relation' without any relation annotations. For example, in the first image, our model is able to give high potential to `wearing' and `has' given the two boxes.

\begin{figure*}[t]
\centering
\begin{minipage}[t]{.45\linewidth}
\includegraphics[width=\linewidth]{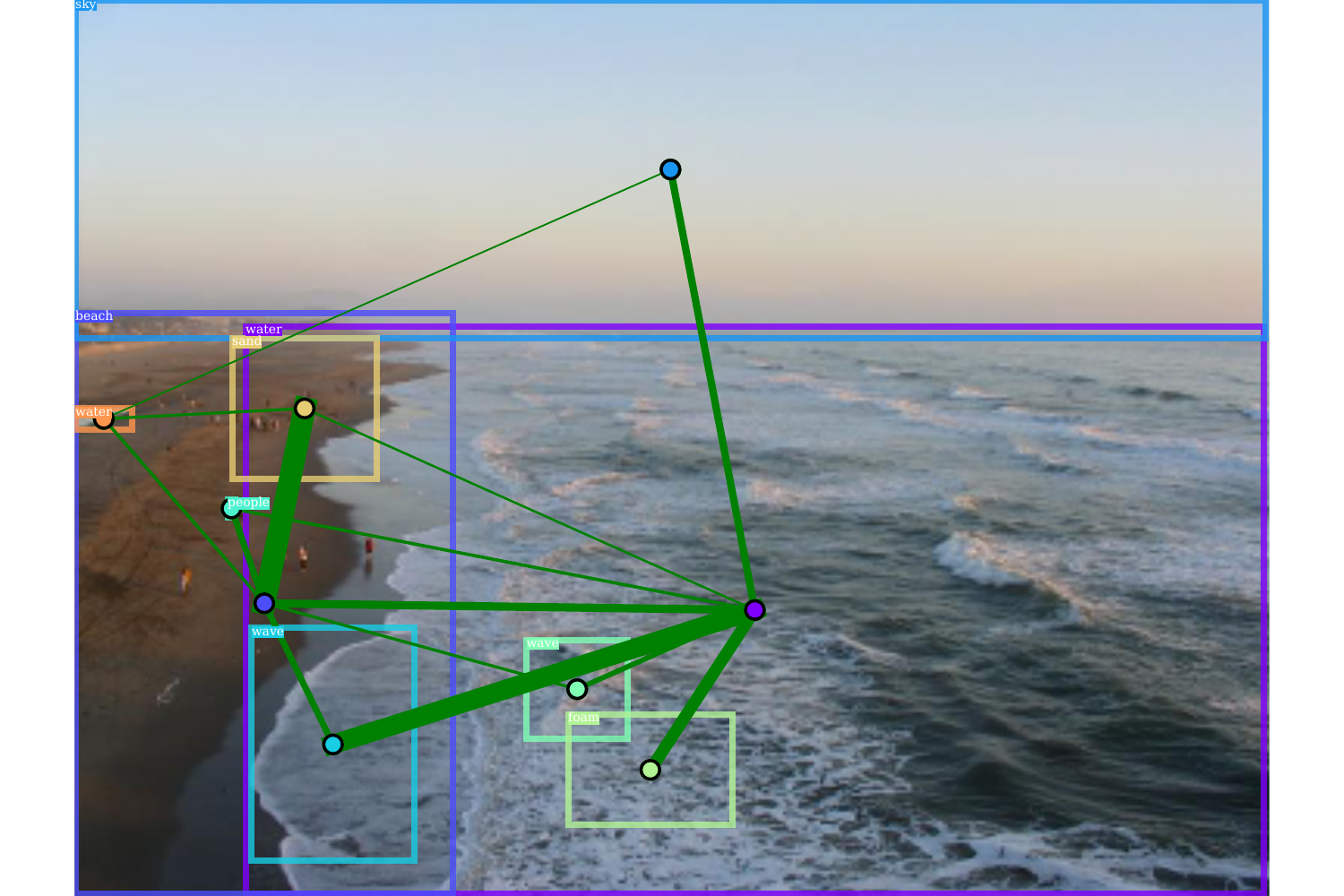}
\end{minipage}
\hspace{.5em}
\begin{minipage}[t]{.45\linewidth}
\includegraphics[width=\linewidth]{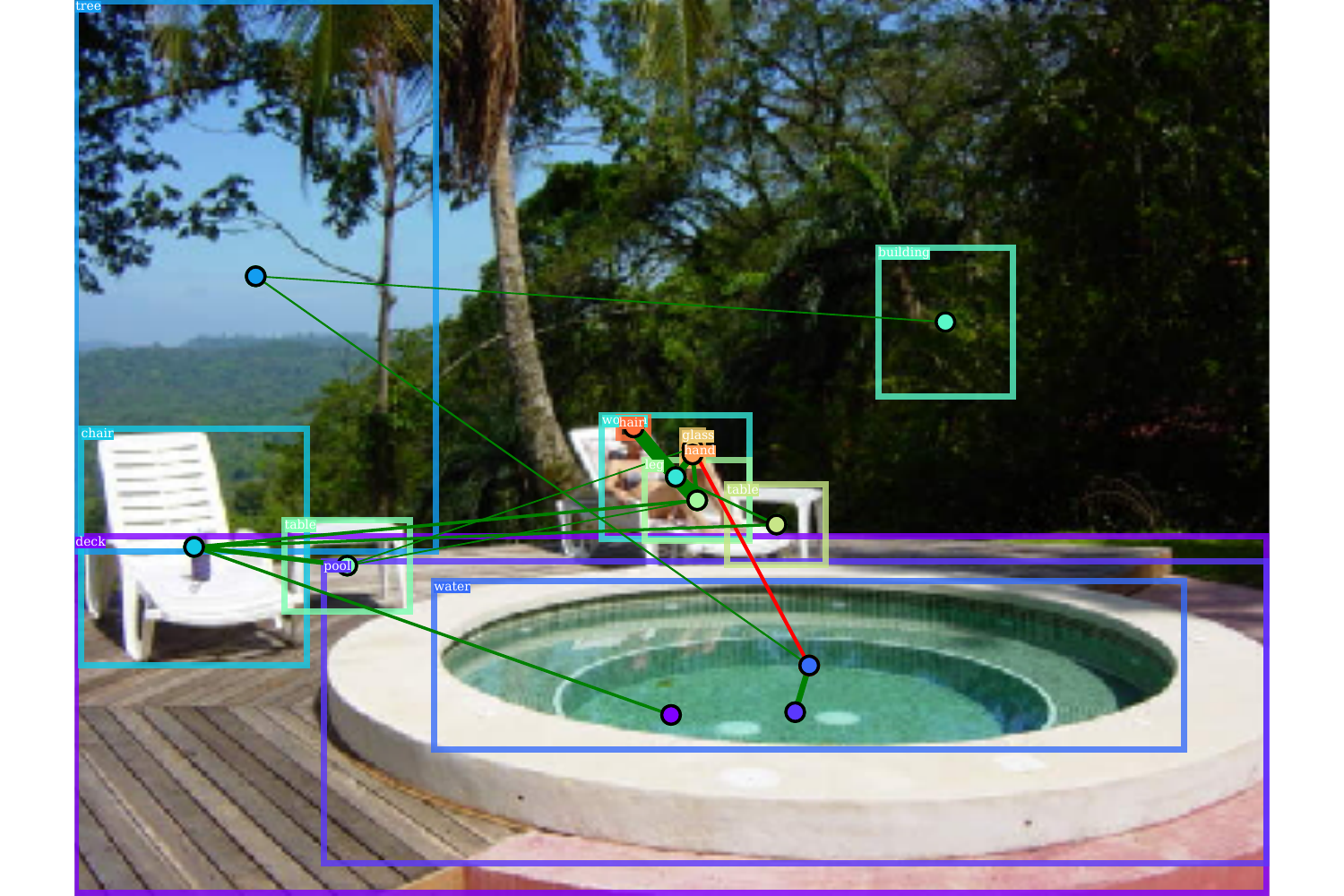}
\end{minipage}
\\ \vspace{0.3cm}
\begin{minipage}[t]{.45\linewidth}
\includegraphics[width=\linewidth]{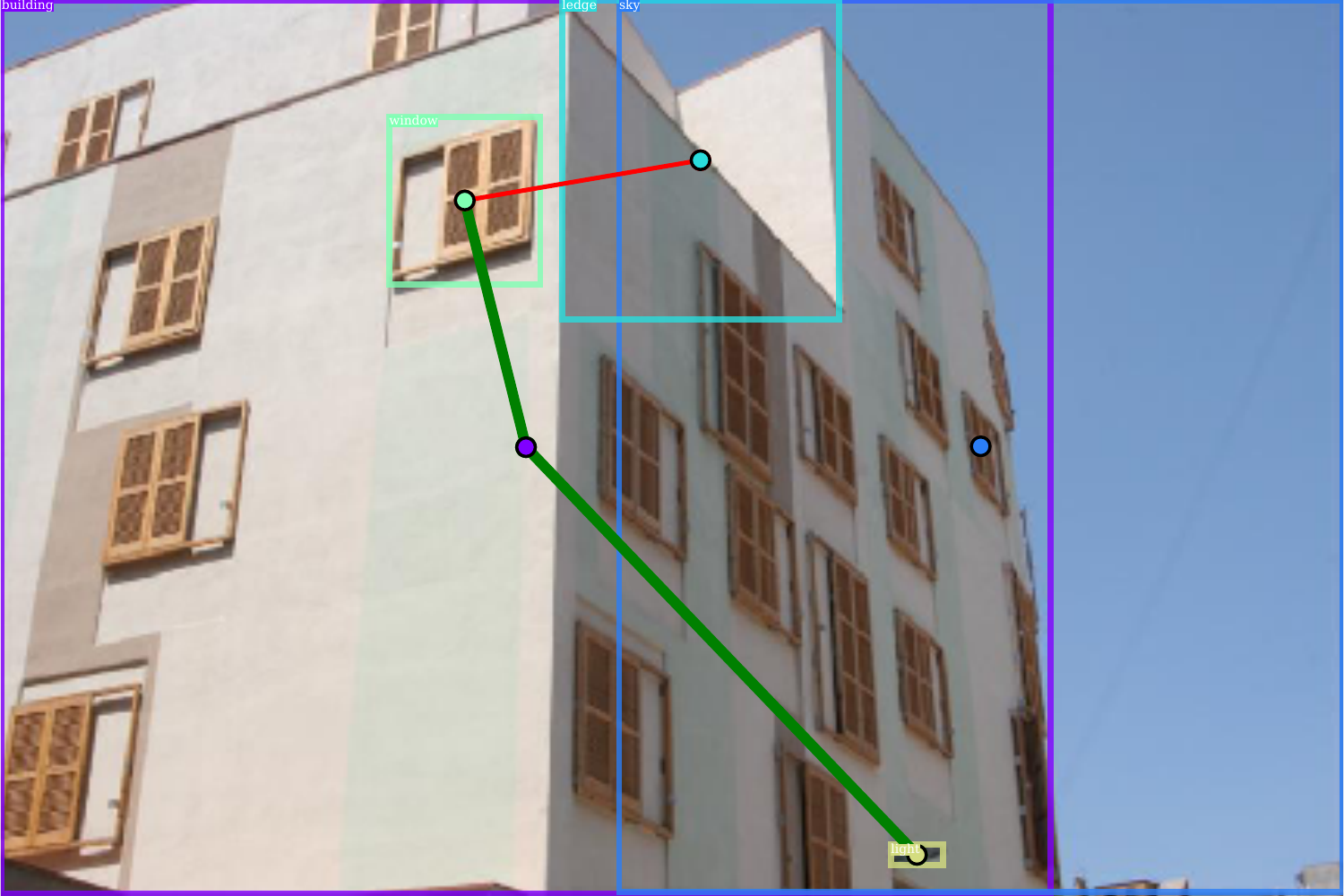}
\end{minipage}
\hspace{.5em}
\begin{minipage}[t]{.45\linewidth}
\includegraphics[width=\linewidth]{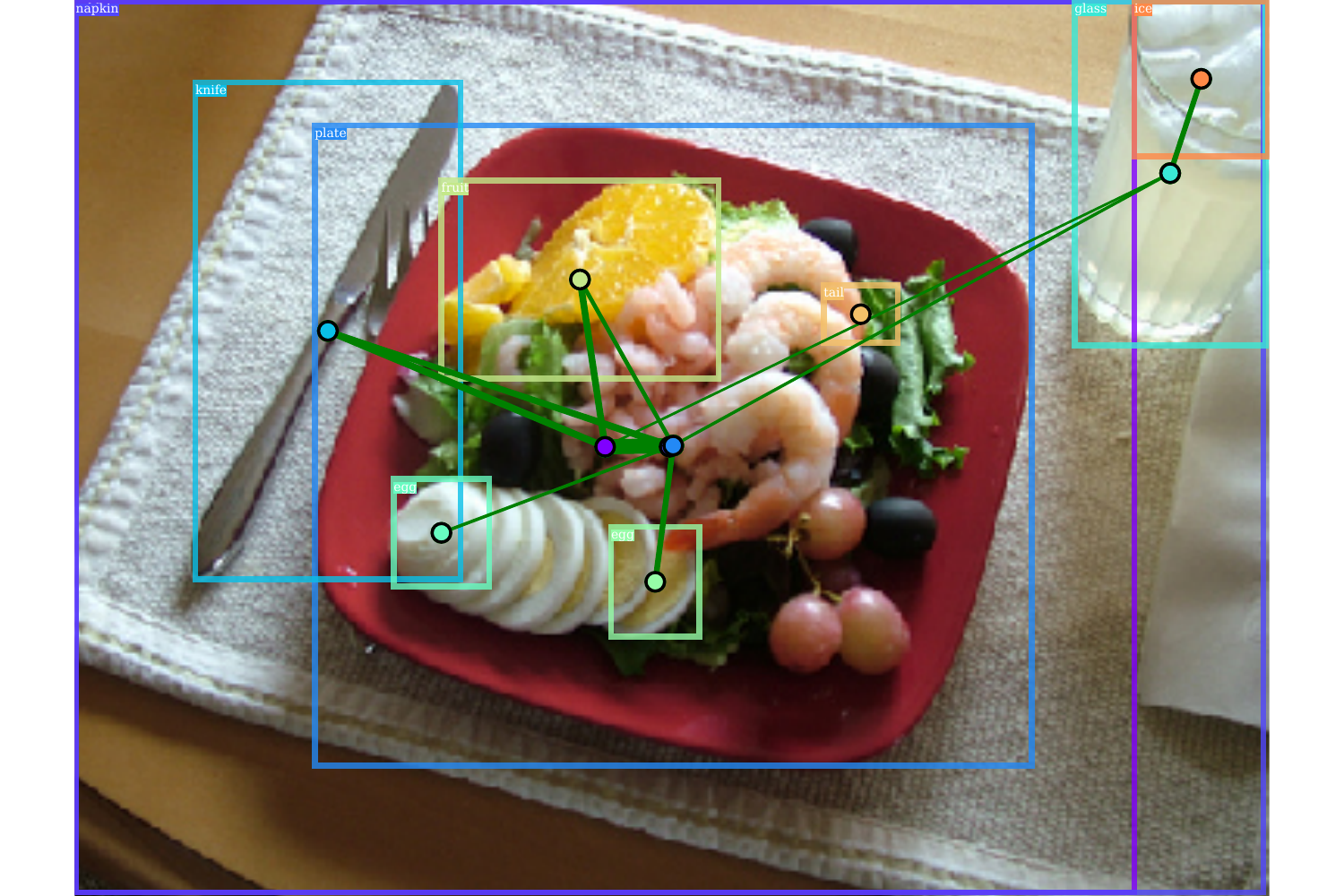}
\end{minipage}
\vspace{0.2cm}
\caption{Visualization of pairwise potentials. Edges with potential less than 0.5 are omitted. The thickness of the line indicates how large the potential is. The ground truth category is annotated on the top-left corner of each box.}
\label{figs:graphs}
\end{figure*}

In Figure~\ref{figs:graphs}, we show the graph in which all the objects in one image are connected by pairwise potentials $\theta_{ij}$. The width of the line is proportional to the corresponding pairwise potential. For better visualization, edges with potential less than 0.5 are omitted. Objects that are related will have higher potentials. For example, in the top-left image, the wave and water has a thick edge since they have a strong relationship given by the pairwise potential.

